\documentclass[default,iicol]{sn-jnl}

\usepackage{graphicx}%
\usepackage{multirow}%
\usepackage{amsmath,amssymb,amsfonts}%
\usepackage{amsthm}%
\usepackage{mathrsfs}%
\usepackage[title]{appendix}%
\usepackage{xcolor}%
\usepackage{textcomp}%
\usepackage{manyfoot}%
\usepackage{booktabs}\let\cline\cmidrule 
\usepackage{algorithm}%
\usepackage{algorithmicx}%
\usepackage{makecell}
\usepackage{algpseudocode}%
\usepackage{listings}%
\usepackage{multicol}
\usepackage{mathtools}
\usepackage{iac_pkg}
\usepackage{lipsum}
\usepackage{amssymb}
\usepackage{pifont}
\usepackage{tabularx, booktabs}
\usepackage[htt]{hyphenat}
\usepackage{xspace}
\usepackage{overpic}
\usepackage{subcaption}
\usepackage{cleveref}

\definecolor{citecolor}{RGB}{34,139,34}
\definecolor{lightred}{RGB}{255,100,100}
\definecolor{cell_bisque}{rgb}{1.0, 0.89, 0.77}
\definecolor{cell_blond}{rgb}{0.98, 0.94, 0.75}
\definecolor{cell_blue}{RGB}{155, 187, 228}
\definecolor{princetonorange}{rgb}{1.0, 0.56, 0.0}
\definecolor{pinkpearl}{rgb}{0.91, 0.67, 0.81}
\definecolor{apricot}{rgb}{0.98, 0.81, 0.69}
\definecolor{mossgreen}{rgb}{0.68, 0.87, 0.68}

\newcommand{\Paragraph}[1]{\noindent\textbf{#1.}}
\newcommand{\Section}[1]{\section{#1}}
\newcommand{\SubSection}[1]{\subsection{#1} }

\begin{document}

\title[Language-guided Hierarchical Fine-grained Image Forgery Detection and Localization]{Language-guided Hierarchical Fine-grained Image Forgery Detection and Localization}


\author[1$^{*}$]{\fnm{Xiao} \sur{Guo}}\email{guoxia11@msu.edu}

\author[2]{\fnm{Xiaohong} \sur{Liu}}\email{xiaohongliu@sjtu.edu.cn}
\author[3]{\fnm{Iacopo} \sur{Masi}}\email{masi@di.uniroma1.it}
\author[1]{\fnm{Xiaoming} \sur{Liu}}\email{liuxm@msu.edu}

\affil[1$^{*}$]{\orgdiv{Department of Computer Science and Engineering}, \orgname{Michigan State University}, \orgaddress{ \country{USA}}}

\affil[2]{\orgdiv{John Hopcroft Center for Computer Science}, \orgname{Shanghai Jiao Tong University}, \orgaddress{ \country{China}}}

\affil[3]{\orgdiv{Department of Computer Science}, \orgname{Sapienza, University of Rome}, \orgaddress{ \country{Italy}}}


\abstract{Differences in forgery attributes of images generated in CNN-synthesized and image-editing domains are large, and such differences make a unified image forgery detection and localization (IFDL) challenging. 
To this end, we present a hierarchical fine-grained formulation for IFDL representation learning. Specifically, we first represent forgery attributes of a manipulated image with multiple labels at different levels. 
Then, we perform fine-grained classification at these levels using the hierarchical dependency between them. 
As a result, the algorithm is encouraged to learn both comprehensive features and the inherent hierarchical nature of different forgery attributes, thereby improving the IFDL representation. 
In this work, we propose a Language-guided Hierarchical Fine-grained IFDL, denoted as HiFi-Net++. 
Specifically, HiFi-Net++ contains four components: a multi-branch feature extractor, a language-guided forgery localization enhancer, as well as classification and localization modules. 
Each branch of the multi-branch feature extractor learns to classify forgery attributes at one level, while localization and classification modules segment the pixel-level forgery region and detect image-level forgery, respectively. 
In addition, the language-guided forgery localization enhancer (LFLE), containing image and text encoders learned by contrastive language-image pre-training (CLIP), is used to further enrich the IFDL representation. 
LFLE takes specifically designed texts and the given image as multi-modal inputs and then generates the visual embedding and manipulation score maps, which are used to further improve HiFi-Net++ manipulation localization performance. 
Lastly, we construct a hierarchical fine-grained dataset to facilitate our study. We demonstrate the effectiveness of our method on $8$ by using different benchmarks for both tasks of IFDL and forgery attribute classification. Our source code and dataset can be found: \href{https://github.com/CHELSEA234/HiFi_IFDL}{github.com/CHELSEA234/HiFi-IFDL}.}

\maketitle

\Section{Introduction}\label{sec:intro}

\begin{figure}[t]
    \centering
    \begin{overpic}[width=0.5\textwidth]{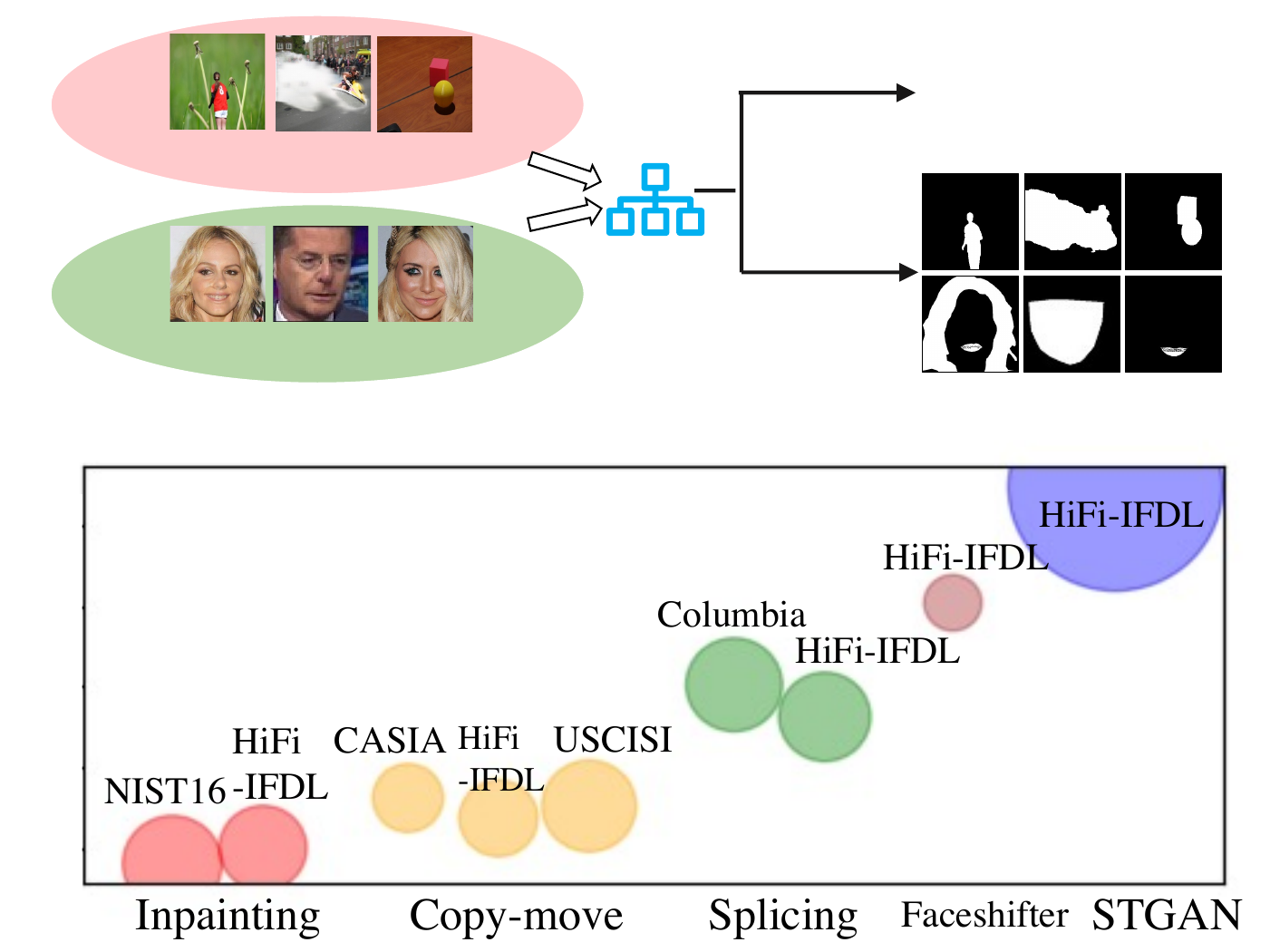}
    \put(55,69){\small{Detection}}
    \put(49,50.5){\small{Localization}}
    \put(72,63.5){\small{\textcolor{citecolor}{Real} \textit{vs} \textcolor{lightred}{Forgery}}}
    \put(12,61.5){\small{Image Editing}}
    \put(10,46.5){\small{CNN-synthesized}}
    \put(49,41){(a)}
    
    \put(0,1){\rotatebox{90}{\tbf{\scriptsize{Mean/Variance of forgery area}}}}
    \put(3.5,33){\footnotesize{$.5$}}
    \put(3.5,26.5){\footnotesize{$.4$}}
    \put(3.5,20){\footnotesize{$.3$}}
    \put(3.5,13.5){\footnotesize{$.2$}}
    \put(3.5,7.5){\footnotesize{$.1$}}
    \put(49,-1.5){(b)}
    \end{overpic}
    \vspace{1mm}
    \caption{(a) In this work, we study image forgery detection and localization (IFDL), regardless of forgery method domains. 
    (b) The distribution of forgery regions depends on individual forgery methods. Each color represents one forgery category (x-axis). Each bubble represents one image forgery dataset. The y-axis denotes the average of the forgery area. The bubble's area is proportional to the variance of the forgery area.}
    \label{fig_mani_overview}
    \vspace{2mm}
\end{figure}

\begin{figure*}[t]
    \centering
    \begin{overpic}[width=1\textwidth]{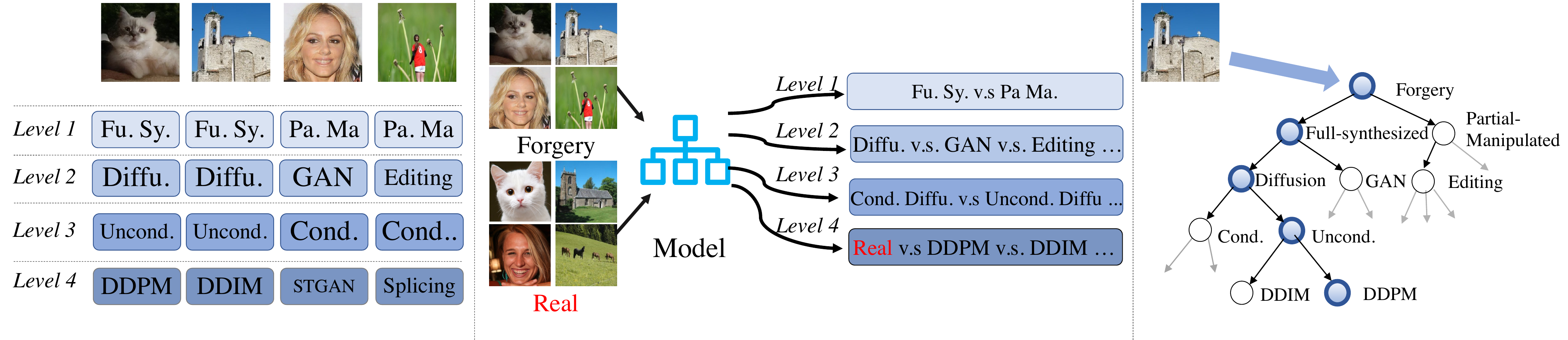}
    \put(18,0){(a)}
    \put(50,0){(b)}
    \put(85,0){(c)}
    \end{overpic}
    \caption{(a) We represent the forgery attribute of each manipulated image with multiple labels at different levels. (b) For an input image, we encourage the algorithm to classify its fine-grained forgery attributes at different levels, \textit{i.e.} a $2$-way classification (fully synthesized or partially manipulated) on level $1$. (c) We perform the fine-grained classification via the hierarchical nature of different forgery attributes, where each depth $l$ node's classification probability is conditioned on classification probabilities of neighbor nodes at depth ($l-1$). [Key: Fu. Sy.: Fully Synthesized; Pa. Ma.: Partially manipulated; Diff.: Diffusion model; Cond.: Conditional; Uncond.: Unconditional].}
    \vspace{-2mm}
    \label{fig_overview_2}
\end{figure*}

Chaotic and pervasive multimedia information sharing offers better means for spreading misinformation~\cite{cnn_internet}, and the forged image content could, in principle, sustain recent ``infodemics''~\cite{infodemics}. Firstly, Convolutional Neural Networks (CNN) synthesized images made extraordinary leaps culminating in recent synthesis methods---Dall$\cdot$E~\cite{ramesh2022hierarchical} or Google ImageN~\cite{saharia2022photorealistic}---based on diffusion models (DDPM)~\cite{ho2020denoising_ddpm}, which even generate realistic videos from text~\cite{singer2022make,ho2022imagen}. 
Secondly, the availability of image editing tool-kits produced substantially low-cost access to image forgery or tampering (\textit{e.g.}, splicing and inpainting). 
In response to such an issue of image forgery, the computer vision community has made considerable efforts, which however branch separately into two directions: detecting either CNN synthesis~\cite{zhangxue2019detecting,wang2020cnn,stehouwer2019detection} or conventional image editing~\cite{wu2019mantra,hu2020span,liu2022pscc,dong2022mvss,wang2022objectformer}. As a result, these methods may be ineffective when deploying to real-life scenarios, where forged images can possibly be generated from either CNN-synthesized or image-editing domains.

To push the frontier of image forensics~\cite{sencar2022multimedia}, we study the image forgery detection and localization problem (IFDL)---Fig.~\ref{fig_mani_overview}\textcolor{blue}{a}---regardless of the forgery method domains, \textit{i.e.}, CNN-synthesized or image editing. 
It is challenging to develop a unified algorithm for two domains, as images generated by different forgery methods differ largely from each other in terms of various forgery attributes. 
For example, a forgery attribute can indicate whether a forged image is fully synthesized or partially manipulated or whether the forgery method used is the diffusion model generating images from the Gaussian noise or an image editing process that splices two images via Poisson editing~\cite{perez2003poisson}. 
Therefore, to model such complex forgery attributes, we first represent the forgery attribute of each forged image with multiple labels at different levels. Then, we present a hierarchical fine-grained formulation for IFDL, which requires the algorithm to classify fine-grained forgery attributes of each image at different levels via the inherent hierarchical nature of different forgery attributes.

Fig.~\ref{fig_overview_2}\textcolor{blue}{a} shows the interpretation of the forgery attribute with a hierarchy, which evolves from the general forgery attribute, fully-synthesized vs partial-manipulated, to specific individual forgery methods, such as DDPM~\cite{ho2020denoising_ddpm} and DDIM~\cite{song2020denoising_ddim}.
Then, given an input image, our method performs fine-grained forgery attribute classification at different levels (see Fig.~\ref{fig_overview_2}\textcolor{blue}{b}). 
The image-level forgery detection benefits from this hierarchy as the fine-grained classification learns the comprehensive IFDL representation to differentiate individual forgery methods. Also, for the pixel-level localization, the fine-grained classification features can serve as a prior to improve the localization. 
This holds since the distribution of the forgery area is prominently correlated with forgery methods, as depicted in Fig.~\ref{fig_mani_overview}\textcolor{blue}{b}.

In Fig.~\ref{fig_overview_2}\textcolor{blue}{c}, we leverage the hierarchical dependency between forgery attributes in fine-grained classification.
Each node's classification probability is conditioned on the path from the root to itself. 
For example, the classification probability at a node of \texttt{DDPM} is conditioned on the classification probability of all nodes in the path of \texttt{Forgery$\rightarrow$ Fully Synthesis$\rightarrow$Diffusion$\rightarrow$Unconditional$\rightarrow$DDPM}. This differs from prior work~\cite{wu2019mantra,marra2018detection,yu2019attributing_image_attribute,marra2019gans}, which assume a ``flat'' structure in which attributes are mutually exclusive. Predicting the entire hierarchical path helps understand forgery attributes from coarse to fine, thereby capturing dependencies among individual forgery attributes.

Therefore, we propose Hierarchical Fine-grained Network (HiFi-Net)~\cite{guo2023hierarchical}, which is the preliminary version of this work and published in Proceeding of the IEEE/CVF Computer Vision and Pattern Recognition (CVPR 2023). Specifically, HiFi-Net has three components: a multi-branch feature extractor, a localization module, and a detection module. 
Each branch of the multi-branch extractor classifies images at one forgery attribute level. 
The localization module generates the forgery mask with the help of a deep-metric learning-based objective, which improves the separation between real and forged pixels. 

\begin{figure*}[t]
    \centering
    \begin{overpic}[width=\textwidth]{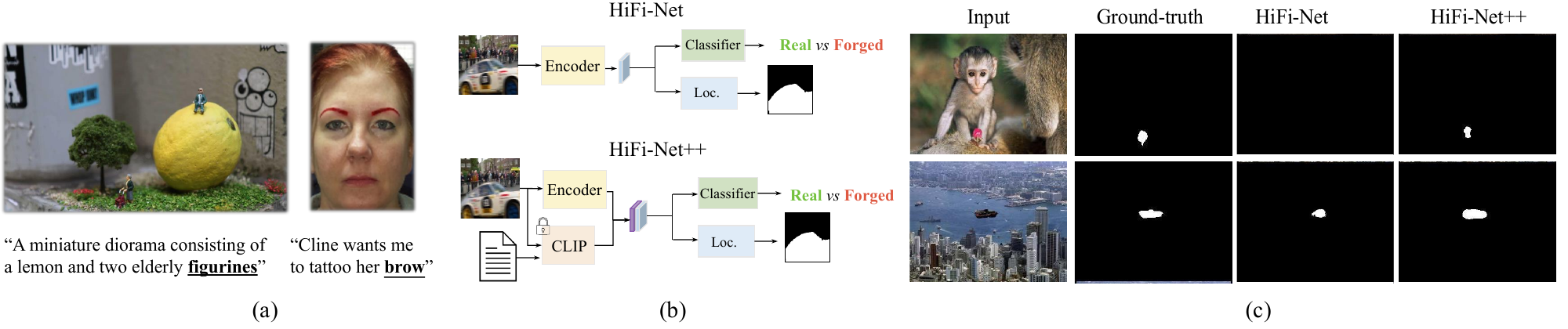}
    \put(44,20){\footnotesize{\cite{guo2023hierarchical}}}
    \put(85,19.7){\footnotesize{\cite{guo2023hierarchical}}}
    \end{overpic}
    \caption{(a) The pre-trained CLIP has a powerful visual embedding that can localize and recognize objects of interest, described by the text query, regardless the objects' spatial size. For example, the pre-trained CLIP can understand the existence of figurines and eyebrows, given the corresponding text query. These images and text queries are acquired via the public CLIP website \href{https://rom1504.github.io/clip-retrieval/}{https://rom1504.github.io/clip-retrieval/}. (b) \textit{Upper}: The HiFi-Net contains three modules, which are multi-branch feature extractor, classification module, and localization module. \textit{Bottom}: We integrated the Language-guided Forgery Localization Enhancer (LFLE) into the existing HiFi-Net, which is then denoted as HiFi-Net++. (c) The HiFi-Net++ can localize manipulation accurately, even when the manipulation area is small. 
    }
    \label{fig:clip_intro}
\end{figure*}

Although the proposed HiFi-Net~\cite{guo2023hierarchical} achieves the state-of-the-art IFDL performance, its performance becomes suboptimal when the manipulation area is small, as detailed in Fig.~9 of the work~\cite{guo2023hierarchical}. Furthermore, it is always desirable to improve the generalization capacity of IFDL solution~\cite{wang2020cnn,ojha2023towards}. Given these two considerations, we empirically observe that leveraging the fixed visual embedding of the pre-trained CLIP image encoder can improve the HiFi-Net's localization performance on images with small manipulation regions and boost the overall generalization ability. We believe this is because the pre-trained CLIP image encoder has two merits. First, CLIP is trained to match images to corresponding texts, which means the pre-trained CLIP visual embeddings excel at locating objects of interest, regardless of the object's spatial size---Fig.~\ref{fig:clip_intro}\textcolor{blue}{(a)}. Secondly, the pre-trained CLIP is exposed to a vast number of web data, and its visual embedding is observed to generalize well for differentiating real and fake images~\cite{ojha2023towards}. The Fig.~2 in work~\cite{wu2023generalizable} further supports this hypothesis.

Also, although the pre-trained CLIP model is trained to maximize the correlation between same text-image vector pairs, the computer vision community recently, in fact, adapts this pre-trained foundation model for text-pixel matching. This adaptation involves leveraging pre-trained CLIP to achieve remarkable outcomes in pixel-wise classification tasks, such as image segmentation~\cite{zhou2022extract,xu2022groupvit,xu2023open,ghiasi2022scaling,li2022languagedriven} and object detection~\cite{zhong2022regionclip,rao2022denseclip}. This success is attributed to the fact that the pre-trained CLIP has a powerful representation ability that matches or grounds the rich semantic text embedding to certain visual contents expressed by the visual embedding produced by the pre-trained CLIP image encoder.

\begin{figure*}[t]
  \centering
    \includegraphics[width=0.95\textwidth]{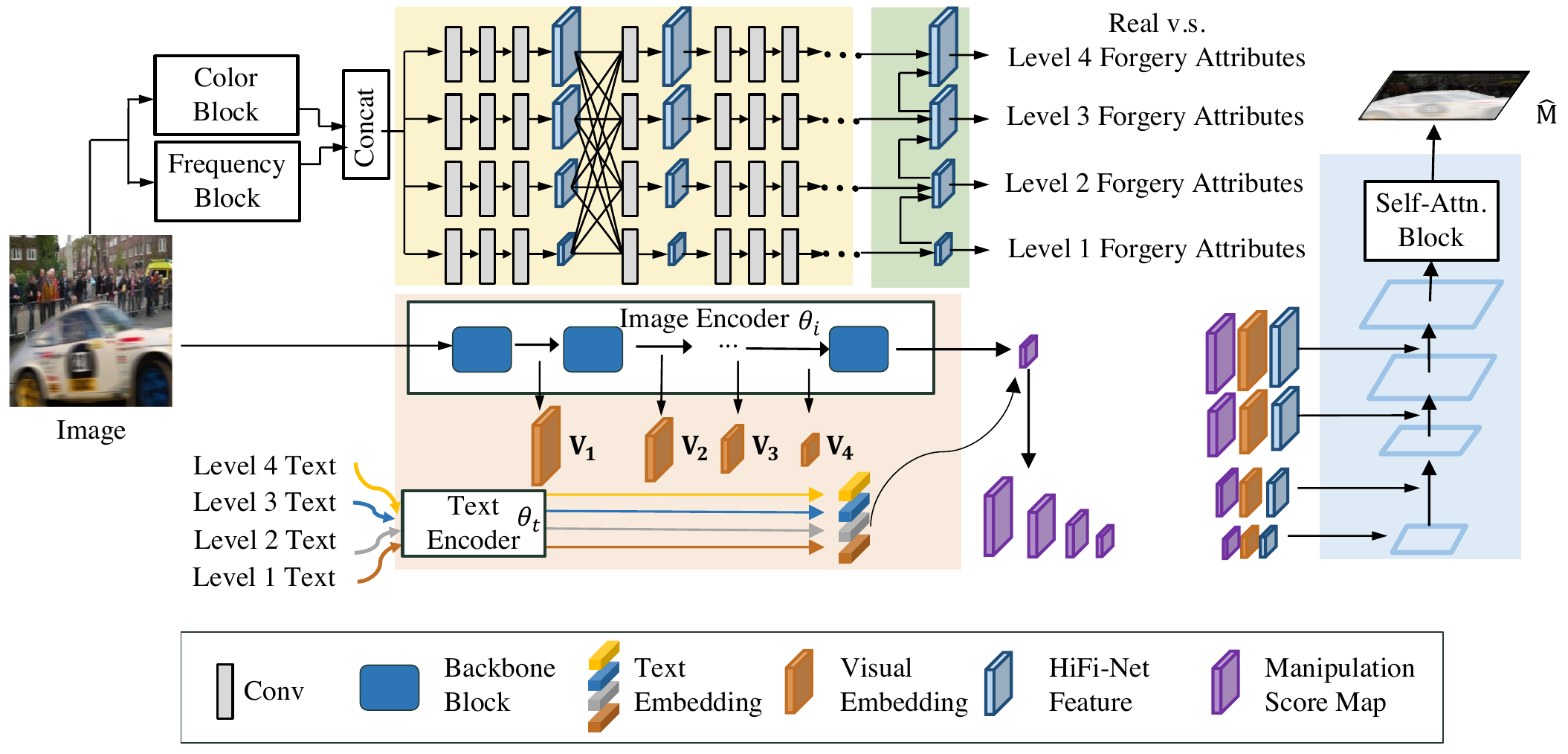}
    \caption{
        Given the input image, we first leverage color and frequency blocks to extract features. The multi-branch feature extractor~(\textcolor{cell_blond}{\rule{0.4cm}{0.25cm}}) learns feature maps of different resolutions, and these feature maps are used for the fine-grained classification at different levels via the classification module (\textcolor{mossgreen}{\rule{0.4cm}{0.25cm}}), detailed in Sec.~\ref{sec_cls}. The language-guided localization enhancer (\textcolor{apricot}{\rule{0.4cm}{0.25cm}}), containing the pre-trained CLIP image and text encoders (denoted as $\net_i$ and $\net_t$ respectively), takes the input image and pre-defined text input, and then produce the visual embedding and the manipulation score map. The entire process is detailed in Sec.~\ref{sec_lang_guided_fe}. In the end, the localization module (\textcolor{cell_blue}{\rule{0.4cm}{0.25cm}}) in Sec.~\ref{sec_loc} jointly takes HiFi-Net feature map, visual embedding, and manipulation score map to generate the binary mask $\hat{\mathbf{M}}$ that indicates the manipulation area.
        }
    \label{fig:architecture}
\end{figure*}

Motivated by the above discussions, we therefore propose HiFi-Net++, which has an additional Language-guided Forgery Localization Enhancer (LFLE) module on the top of the HiFi-Net~\cite{guo2023hierarchical}, as shown in Fig.~\ref{fig:clip_intro}\textcolor{blue}{(b)}. This LFLE module consists of the pre-trained CLIP image and text encoder, as well as a refinement block that helps optimize the text embedding to better adapt to the current IFDL task. 
Specifically, given the input image, the pre-trained CLIP image encoder provides the \emph{visual embedding}. After that, we use specifically designed templates with the name of each aforementioned forgery attribute (\textit{e.g.}, \texttt{fully-synthesized}, \texttt{partial-manipulated}, \texttt{GAN}, etc.) to construct text inputs, and then apply the pre-trained CLIP text encoder on these text inputs to obtain corresponding text embeddings. Such text embeddings are used with the visual features output from the pre-trained CLIP image encoder to generate a 2D manipulation score map. This manipulation score map denotes the spatial location manipulated by a given forgery method.
As depicted in Fig.~\ref{fig:architecture}, the highly generalizable visual embedding and manipulation score maps are used altogether to help localize the manipulation.
It is worth noting that, being inspired by the prior works~\cite{zhou2022learning,gao2023clip,zhang2021tip,yao2021cpt}, we also use a refinement block to enhance the text embedding, such that we can adapt the pre-trained CLIP text encoder and its generated text embedding for our manipulation localization tasks.

Lastly, to facilitate our study of the hierarchical fine-grained formulation, we construct a new dataset, termed the Hierarchical Fine-grained (HiFi) IFDL dataset. It contains $13$ forgery methods, which are either the latest CNN-synthesized methods or representative image editing methods. HiFi-IFDL dataset also induces a hierarchical structure on forgery categories to enable learning a classifier for various forgery attributes.
Each forged image is also paired with a high-resolution ground truth forgery mask for the localization task.

In summary, we extend the preliminary version of this work~\cite{guo2023hierarchical} in two aspects: (i) by utilizing the fixed visual embedding from the pre-trained CLIP image encoder, the HiFi-Net++ has improved the localization performance of spatially small manipulations as well as the generalization ability to unseen manipulation; (ii) we adapt the idea of text-pixel matching for the manipulation localization task --- use the pre-trained CLIP text embeddings with visual features to produce the manipulation score map, which serves as an auxiliary signal to help the localization performance.

The contributions of this work are as follows:

$\diamond$ We study the task of image forgery detection and localization (IFDL) for both image editing and CNN-synthesized domains. We propose a hierarchical fine-grained formulation to learn a comprehensive representation of IFDL and forgery attribute classification.

$\diamond$ We propose an IFDL algorithm named HiFi-Net, which not only performs well on forgery detection and localization but also identifies a diverse spectrum of forgery attributes. Based on this proposed HiFi-Net in the preliminary version of this work, we further introduce a more advanced version IFDL algorithm, namely HiFi-Net++. HiFi-Net++ utilizes an additional language-guided forgery localization enhancer (LFLE) to improve the manipulation localization performance and its generalization ability.

$\diamond$ We construct a new dataset (HiFi-IFDL) to facilitate the hierarchical fine-grained IFDL study. When evaluating over $8$ benchmarks, our method outperforms state-of-the-art (SoTA) on the tasks of IFDL and achieves competitive performance on the forgery attribute classifications.


\begin{table*}[t]
\centering
\resizebox{0.98\linewidth}{!}{
\begin{tabular}{@{}ccccccc@{}}
    \toprule
    \tbf{Method} & {Detection} & {Localization} & \shortstack[l]{Forgery\\ Type} & \shortstack[l]{Attribute\\ Learning} & \shortstack[l]{Fixed CLIP\\ Visual Embed.} & \shortstack[l]{Text-pixel\\Matching}\\ \midrule\midrule
    Wu \etal~\cite{wu2019mantra} & \redcross & \greenv & Editing & \redcross & N/A & N/A\\ 
    Hu \etal~\cite{hu2020span} & \redcross & \greenv & {Editing} & \redcross & N/A & N/A\\ 
    Liu \etal~\cite{liu2022pscc} & \greenv & \greenv & Editing & \redcross& N/A & N/A\\ 
    Dong \etal~\cite{dong2022mvss} & \greenv & \greenv & Editing & \redcross& N/A & N/A\\ 
    Wang \etal~\cite{wang2022objectformer} & \greenv & \greenv & Editing & \redcross& N/A & N/A\\ 
    Zhang \etal~\cite{zhangxue2019detecting} & \greenv & \redcross & CNN-based & \redcross&N/A & N/A\\ 
    Wang \etal~\cite{wang2020cnn} & \greenv& \redcross & CNN-based & \redcross& N/A & N/A\\ 
    Asnani \etal~\cite{asnani2021reverse} & \greenv & \redcross & CNN-based & syn.-based& N/A & N/A\\ 
    Yu \etal~\cite{yu2019attributing_image_attribute} & \greenv & \redcross & CNN-based & syn.-based& N/A & N/A\\ 
    Stehouwer \etal~\cite{stehouwer2019detection} & \greenv & \greenv & CNN-based & \redcross& N/A & N/A\\ 
    Huang \etal~\cite{huang2020fakelocator} & 
    \greenv & \greenv & CNN-based & \redcross& N/A & N/A\\ 
    Guo \etal~\cite{guo2023hierarchical} & \greenv & \greenv & Both & \greenv & N/A & N/A\\ 
    Wu \etal~\cite{wu2023generalizable}  & 
    \greenv & \redcross & CNN-based & \redcross & \redcross& \redcross\\
    Sun \etal~\cite{sun2023towards} & \greenv & \redcross & CNN-based & \redcross & \redcross & \redcross \\
    Ojha \etal~\cite{ojha2023towards} & \greenv & \redcross & CNN-based & \redcross & \greenv & \redcross \\
    Sha \etal~\cite{sha2023de-fake} & \greenv & \redcross & CNN-based & \redcross & \greenv & \redcross \\
    Rao \etal~\cite{rao2022denseclip} & N/A & N/A & N/A & N/A & \redcross & \greenv \\
    Zhou \etal~\cite{zhou2022extract} & N/A & N/A & N/A & N/A & \greenv & \greenv \\
    Xu \etal~\cite{xu2022groupvit} & N/A & N/A & N/A & N/A & \redcross & \greenv \\
    Ghiasi \etal~\cite{ghiasi2022scaling} & N/A & N/A & N/A & N/A & \redcross & \greenv \\
    \midrule\midrule
    \
    \tbf{HiFi-Net++} & \greenv & \greenv & Both & \greenv & \greenv & \greenv\\
    \bottomrule
\end{tabular}}
\caption{Comparison to previous works.}
\label{tab_background_compare}
\end{table*}

\Section{Related Work}\label{sec:related}
\Paragraph{Image Forgery Detection and Localization} 
In the generic image forgery detection, it is required to distinguish real images from ones generated by a CNN: Zhang \etal \cite{zhangxue2019detecting} report that it is difficult for classifiers to generalize across different GANs and leverage upsampling artifacts as a strong discriminator for GAN detection. 
On the contrary, against expectation, the work by Wang \etal~\cite{wang2020cnn} shows that a baseline classifier \emph{can} actually generalize in detecting different GAN models contingent to being trained on synthesized images from ProGAN~\cite{karras2018progressive}.
Another thread is facial forgery detection~\cite{KorshunovICB2019,rossler2019faceforensics++v3,google_dfd,dolhansky_deepfake_2019,li_2020_CVPR,Jiang_2020_CVPR,unified-detection-of-digital-and-physical-face-attacks,RED_journal,zhang2024common,guo2023tracing}. 
Notably, recent DD-VQA~\cite{zhang2024common} introduces a novel approach by generating interpretative explanations for forgery detection, significantly enhancing the interpretability by not only identifying the forgeries but also providing clear, human-understandable explanations behind the model's decision.
MM-Det~\cite{xiufeng_lamma_detection} makes the first attempt to leverage a large multi-modal model to discern unseen forgeries, achieving remarkable detection performance for diffusion model generated contents.  
All these works specialize in image-level forgery detection, which, however, does not meet the need of knowing where the forgery occurs on the pixel level. 
Therefore, we perform both image forgery detection and localization, as reported in Tab.~\ref{tab_background_compare}.

As for the forgery localization, most existing methods perform pixel-wise classification to identify forged regions~\cite{wu2019mantra,hu2020span,wang2022objectformer} while early ones use a region~\cite{zhou2018learning} or patch-based~\cite{mayer2018learned} approach. The idea of localizing forgery is also adopted in the DeepFake Detection community by segmenting the artifacts in facial images~\cite{zhao2021learning,chai2020makes,cozzolino2018forensictransfer}. 
Zhou \etal~\cite{zhou2020generate} improve the localization by focusing on object boundary artifacts. 
The MVSS-Net~\cite{chen2021image,dong2022mvss} uses multi-level supervision 
to balance between sensitivity and specificity. 

\Paragraph{Attribute Learning} CNN-synthesized image attributes can be observed in the frequency domain~\cite{zhangxue2019detecting,wang2020cnn}, where different GAN generation methods have distinct high-frequency patterns. 
The task of ``GAN discovery and attribution'' attempts to identify the exact generative model~\cite{marra2018detection,yu2019attributing_image_attribute,marra2019gans,pan2024towards}.
These works differ from ours in two aspects. 
Firstly, the prior work concentrates on the attribute used in the digital synthesis method (synthesis-based), yet our work studies forgery-based attribute, \textit{i.e.}, to classify GAN-based fully-synthesized or partial manipulation from the image editing process. 
Secondly, unlike the prior work that assumes a ``flat'' structure between different attributes, we represent all forgery attributes in a hierarchical way, exploring dependencies among them.

\Paragraph{CLIP in Image Understanding} 
Recently, CLIP~\cite{radford2021learning} is employed by the computer vision community for core vision tasks, such as image classification, segmentation~\cite{zhou2022extract,xu2022groupvit,xu2023open,ghiasi2022scaling,li2022languagedriven}, object detection~\cite{zhong2022regionclip,rao2022denseclip}, and scene understanding~\cite{zhang2024tamm,zhang2024vision,zhang2024spartun3d,zhang2024navhint,zhang2023-vl-trans}, etc. 
In this work, we are interested in two topics: (1) CLIP-based image forensic methods and (2) CLIP-based pixel-matching work.

For CLIP-based image forensic methods, Uni-Det~\cite{ojha2023towards} 
applies $k$-NN and linear probing on the fixed pre-trained CLIP visual embedding.
This approach demonstrates remarkable image-level detection performance that generalizes well to many unseen forgery types.
DE-FAKE~\cite{sha2023de-fake} leverages pre-trained CLIP visual and textual embeddings to effectively identify images generated from text-to-image diffusion models.
However, these prior methods have two major differences from our proposed HiFi-Net++. 
First, prior works apply simple architectures (\textit{e.g.}, ResNet$18$, $2$ MLP layers, etc.) on pre-trained CLIP visual embeddings. In contrast, we meticulously devise a multi-branch feature extractor that comprehensively learns forgery attributes across multiple scales and hierarchical relations among these forgery attributes, empirically resulting in better image-level forgery detection performance. 
Secondly, prior works can only perform image-level forgery detection, but our method not only detects image-level forgeries but also localizes manipulated pixels, which is a pixel-level classification task. 

On the other hand, the pre-trained CLIP can also be effective on the pixel-wise vision tasks:~\cite{zhou2022extract} offers a study that finetunes CLIP for semantic segmentation.
\cite{ghiasi2022scaling} points out that CLIP and ALIGN~\cite{jia2021scaling} models can also roughly learn to group objects and are not able to perform localization. \cite{li2022languagedriven} introduces LSeg, a language-driven semantic image segmentation method that replicates CLIP contrastive training at the pixel level and spatial regularization blocks. 
MaskCLIP+~\cite{dong2023maskclip} has been recently outperformed by CLIP-S$^4$~\cite{he2023clip}, not requiring unknown classes to be specified in the training phase. 
Other very recent works are~\cite{xu2023open}, performing open-vocabulary, panoptic segmentation and
DenseCLIP~\cite{rao2022denseclip} that uses the CLIP to serve as the auxiliary input signal to enhance dense prediction. Finally, \cite{xu2022groupvit} proposes a hierarchical Grouping Vision Transformer for image semantic segmentation supervised only by natural language. 
Please note that, unlike MaskCLIP+, which uses pre-trained CLIP text embedding to segment common objects (\textit{e.g.}, dog, cyclist, and tree), we are working on more challenging tasks that segment manipulated pixels by a forgery method. Therefore, we use multi-head attention to refine the pre-trained text embedding for our image forensic tasks.


\begin{figure*}[t]
    \centering
    \includegraphics[width=1\textwidth]{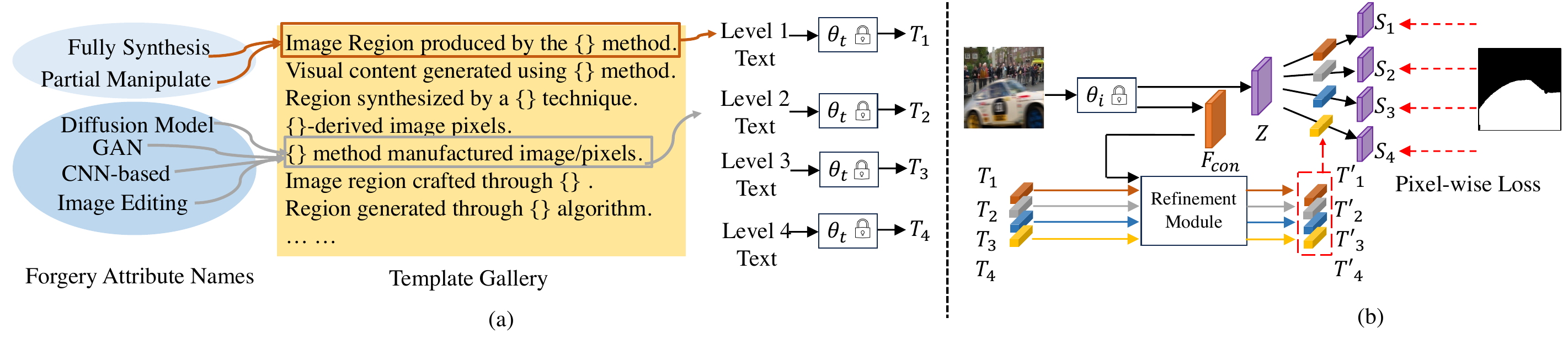}
    \caption{(a) Two forgery attribute names on level 1 of the hierarchical fine-grained formulation (Fig.~\ref{fig_taxonomy_benchmark}) are \texttt{Fully Synthesis} and \texttt{Partial Manipulation}. We combine these forgery attribute names with a template (\textit{e.g.}, \texttt{"Image region is synthesized by \{\} method”}), which is randomly chosen from the template gallery. Therefore, the level 1 text input has two sentences: \texttt{"image is manipulated by fully-synthesized method"}, \texttt{"image is manipulated by partial-manipulated methods"}. Consequently, the level 2, level 3 and level 4 have $4$, $6$, and $13$ sentences as the text input. We apply the pre-trained CLIP text encoder (\textit{e.g.}, $\net_{t}$) on text inputs at different levels and obtain text embedding $\mathbf{T}_{1}$, $\mathbf{T}_{2}$, $\mathbf{T}_{3}$ and $\mathbf{T}_{4}$. (b) The pre-trained CLIP image encoder (\textit{e.g.}, $\net_{i}$) takes the input image, we obtain $\mathbf{F}_{con}$ to represent the visual content, and feature map $\mathbf{Z}$ that maintains the ability to be aligned with text embedding. After that, a refinement module utilizes $\mathbf{F}_{con}$ to conduct refinements on text embeddings at different levels. These refined text embedding ($\mathbf{T}^{\prime}_{b} \text{ with } b \in \{1 \ldots 4\}$) along with $\mathbf{Z}$ generates manipulation score map $\mathbf{S}_{b} \text{ with } b \in \{1 \ldots 4\}$, as the auxiliary signal to help the localization. During the training, we only keep the refinement module as trainable, while pre-trained CLIP image and text encoders are frozen.
    }
    \label{fig:clip_module}
\end{figure*}

\Section{HiFi-Net++}\label{sec:method}

In this section, we introduce HiFi-Net++ (as shown in Fig.~\ref{fig:architecture}) and highlight the modifications made to the original model presented in~\cite{guo2023hierarchical}. We start by defining the image forgery detection and localization (IFDL) task and hierarchical fine-grained formulation. 
In IFDL, an image $\mbf{X} \in \takeval{R}{0}{255}^{W \PLH H \PLH 3}$ needs to be mapped to a binary variable $\mbf{y}$ to predict if it has been forged or is a pristine image. The method also can perform localization of the manipulation at the pixel level, thereby performing binary segmentation and outputting a binary mask $\mbf{M} \in \takeval{R}{0}{1}^{W \PLH H}$, where the $\mbf{M}_{ij}$ indicates if the $ij$-th pixel has been manipulated or not.

In the hierarchical fine-grained formulation, we train the given IFDL algorithm towards fine-grained classifications, and in the inference we evaluate the binary classification results on the image-level forgery detection. Specifically, we denote a categorical variable $\hat{\mbf{y}}_b$ at branch $b$, where its value depends on which level we conduct the fine-grained forgery attribute classification. 
For example, as depicted in Fig.~\ref{fig_overview_2}\textcolor{blue}{(b)}, two forgery attribute categories at level $1$ are full-synthesized, partial-manipulated; four categories at level $2$ are diffusion model, GAN-based method, image editing, CNN-based partial-manipulated method; categories at level $3$ discriminate whether forgery methods are conditional or unconditional; $14$ classes at level $4$ are real and $13$ specific forgery methods.
We detail this in Sec.~\ref{sec_benchmark} and Fig.~\ref{fig_taxonomy_benchmark}. 

In our preliminary work, the proposed HiFi-Net consists of a multi-branch feature extractor (Sec.~\ref{sec_multi_scale}) that performs fine-grained classifications at different specific forgery attribute levels and two modules (Sec.~\ref{sec_cls} and Sec.~\ref{sec_loc}) that help the forgery detection and localization, respectively. In the HiFi-Net++, we introduce an additional Language-guided Forgery Localization Enhancer (IFLE) on the top of the HiFi-Net. IFLE leverages the pre-trained CLIP~\cite{radford2021learning} image encoder $\net_i$ and text encoder $\net_t$ to improve manipulation localization performance and the model's generalization ability. The details are reported in Sec.~\ref{sec_lang_guided_fe}.
\SubSection{Multi-Branch Feature Extractor} \label{sec_multi_scale}
We first extract the feature of the given input image via the color and frequency blocks, and this frequency block applies a Laplacian of Gaussian (LoG)~\cite{burt1987laplacian_laplacia_operator} onto the CNN feature map. This architecture design is similar to the method in~\cite{masi2020two}, which exploits image generation artifacts that can exist in both RGB and frequency domain~\cite{wang2022objectformer,dong2022mvss,wang2020cnn,zhangxue2019detecting}.

Then, we propose a multi-branch feature extractor, whose branch is denoted as $\net_b$ with $b \in \{1 \ldots 4\}$. Specifically, each $\net_b$ generates the feature map of a specific resolution, and such a feature map helps $\net_b$ conduct the fine-grained classification at the corresponding level. For example, for the finest level (\textit{i.e.}~identifying the individual forgery methods), one needs to model contents at all spatial locations, which requires a high-resolution feature map.
In contrast, it is reasonable to have low-resolution feature maps for the coarsest level (\textit{i.e.}~binary) classification. 

We observe that different forgery methods generate manipulated areas with different distributions (Fig.~\ref{fig_mani_overview}\textcolor{blue}{b}), and different patterns, \textit{e.g.}, deepfake methods~\cite{rossler2019faceforensics++v3,li2019faceshifter} manipulate the whole inner part of the face, whereas STGAN~\cite{liu2019stgan} changes sparse facial attributes such as mouth and eyes.
Therefore, we believe that features used for fine-grained classification can serve as a prior for localization. 
It is important to have such a design for localizing both manipulated images with CNNs or classic image editing.

\SubSection{Classification Module}\label{sec_cls}

\begin{figure}[t]
    \centering
    \begin{overpic}[width=0.95\linewidth]{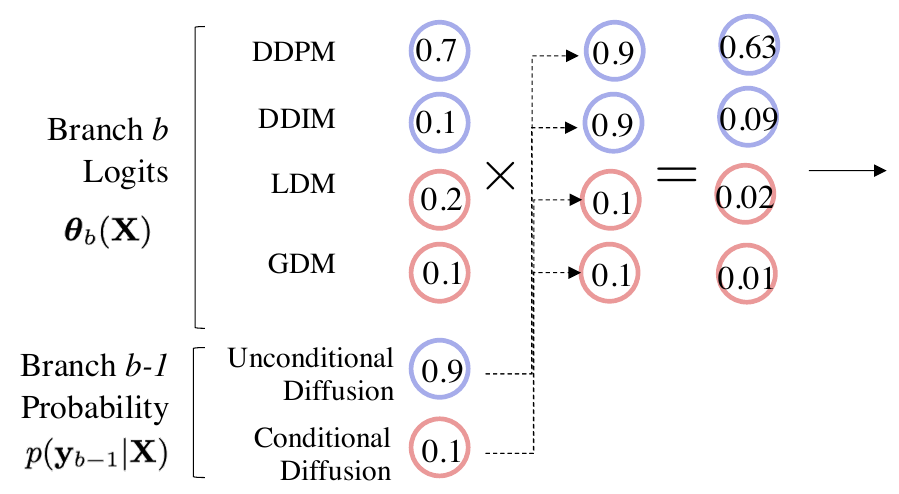}
    \put(37.5,48.5){\small{\cite{ho2020denoising_ddpm}}}
    \put(37.5,41){\small{\cite{song2020denoising_ddim}}}
    \put(37.5,33){\small{\cite{rombach2021highresolution_latent_diffusion}}}
    \put(37.5,24){\small{\cite{nichol2021glide}}}
    \end{overpic}
    \caption{The classification probability output from branch $\net_{b}$ depends on the predicted probability at branch $\net_{b-1}$, following the definition of the hierarchical forgery attributes tree.}\label{fig_path_prediction}
    \vspace{-3mm}
\end{figure}
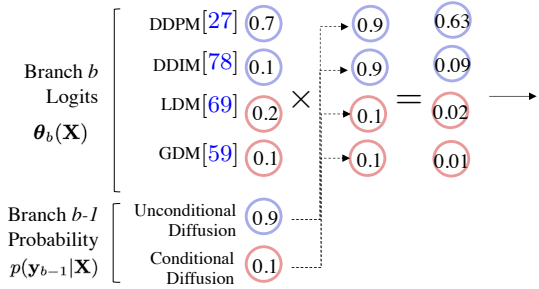

\Paragraph{Hierarchical Path Prediction}
We intend to learn the hierarchical dependency between different forgery attributes. Given the image $\mbf{X}$, we denote output logits and predicted probability of the branch $\net_{b}$ as $\net_b(\mbf{X})$ and $p(\mbf{y}_b|\mbf{X})$, respectively. Then, we have: 
\begin{equation}
    \small
    \begin{split}
    p(\mbf{y}_b|\mbf{X}) \doteq \operatorname{softmax}\Big(\net_b(\mbf{X})\odot (1 + p(\mbf{y}_{b-1}|\mbf{X}))\Big) \quad
    \end{split}
\label{eq_conditional_prob}
\end{equation}
Before computing the probability $p(\mbf{y}_b|\mbf{X})$ at branch $\net_{b}$, we scale logits $\net_b(\mbf{X})$ based on the previous branch probability $p(\mbf{y}_{b-1}|\mbf{X})$. 
Then, we enforce the algorithm to learn hierarchical dependency. Specifically, in Eq.~(\ref{eq_conditional_prob}), we repeat the probability of the coarse level $b-1$ for all the logits output by the branch at level $b$, following the hierarchical structure. Fig.~\ref{fig_path_prediction} shows that the logits associated with predicting \texttt{DDPM} or \texttt{DDIM} are multiplied by probability for the image to be \texttt{Unconditional (Diffusion)} in the last level, according to the tree structure in the hierarchical fine-grained formulation.
\SubSection{Language-guided Forgery Localization Enhancer}
\label{sec_lang_guided_fe}
The overall Language-guided Forgery Localization Enhancer module contains the pre-trained CLIP image $\net_i$ and text encoders $\net_t$, as well as a refinement block that makes the generated text embeddings better adapted to the manipulation localization task.

\Paragraph{Text Input Construction} 
As depicted in Fig.~\ref{fig:clip_module}\textcolor{blue}{(a)}, we first use the name of forgery attributes and pre-defined templates to create text inputs. 
To improve the variability of the text input, we form a template gallery with many text templates generated by a publicly available Large Language Model (LLM)~\cite{ouyang2022training}. Also, we empirically use multiple alternative names for each forgery attribute, for example, \texttt{"GAN"}, \texttt{"Generative Adversarial Networks"} are the alternative names of the same manipulation method (\textit{e.g.}, forgery attribute).
After that, the pre-trained text encoder $\net_t$ takes text inputs to produce the text embeddings.

Formally, we denote the text embedding at different forgery levels as: $\mathbf{T}_1\in \mathbb{R}^{2 \times C}$, $\mathbf{T}_2\in \mathbb{R}^{4 \times C}$, $\mathbf{T}_3\in \mathbb{R}^{6 \times C}$ and $\mathbf{T}_4\in \mathbb{R}^{13 \times C}$, where $\mathbf{T}_b \text{ with } b \in \{1 \ldots 4\}$ indicates the text embedding of the prompt generated at level $b$ and $C$ is the dimension of the text embedding.

\Paragraph{Architecture} 
When the pre-trained CLIP image encoder $\net_i$ takes the given image as input, we obtain $b$ intermediate features maps, denoted as $\mathbf{V}_{b} \in \mathbb{R}^{W_{b} \times H_{b} \times C} \text{ with } b \in \{1 \ldots 4\}$. These feature maps are visual embeddings that are depicted in Fig.~\ref{fig:architecture}, and we empirically find them have a high generalization ability in distinguishing real and manipulated pixels. More details can be found in Tab.~\ref{table_tax_loc} and Fig.~\ref{fig_viz_hifi_editing}. After that, we concatenate these feature maps into $\mathbf{F}_{con}$. Given the powerful representation ability of the pre-trained CLIP model, $\mathbf{F}_{con}$ can contain rich information for visual semantics.
Secondly, we extract the feature map before the final attention pooling in $\net_i$, and denote this feature map as $\mathbf{Z} \in \mathbb{R}^{W\times H\times C}$. In general, in the pre-trained CLIP image encoder, the feature after the final attention pooling is used to correlate with language. However, as indicated in DenseCLIP work~\cite{rao2022denseclip}, even $\mathbf{Z}$ also has the ability to correlate or align with the given text input.

Similar to the previous method that refines the fixed pre-trained CLIP text embedding for specific downstream tasks, we apply a refinement module, which is based on the multi-head attention mechanism. Formally, such a refinement module is applied on the $\mathbf{F}_{con}$ and $\mathbf{T}_{b} \text{ with } b \in \{1 \ldots 4\}$, and then produces a text embedding perturbation $\Delta \mathbf{T}_{b} \text{ with } b \in \{1 \ldots 4\}$
\begin{equation}
    \Delta \mathbf{T}_{b} = \texttt{RefineModule} (\mathbf{T}_{b}, \mathbf{F}_{con}).
\end{equation}
This text embedding perturbation is finally added on the original input text embedding $\mathbf{T}_{b}$ to produce the final text embedding for detection and localization performance as:
\begin{equation}
    \mathbf{T}^{\prime}_{b} = \mathbf{T}_{b} + \Delta \mathbf{T}_{b}.
\end{equation}
We then compute the cosine similarity between each refined text embedding and $\mathbf{Z}$ to yield a manipulation score map as:
\begin{equation}
    \mbf{s}_b \doteq \mathbf{T}^{\prime}_{b}~\mathbf{Z}^{\top} \quad  b \in \{1 \ldots 4\}.
    \label{eq_score_map}
\end{equation}
As a result, this operation yields four manipulation score maps of the spatial dimensionality:
$\mbf{s}_{1}\in \mathbb{R}^{2 \times W \times H}$, $\mbf{s}_{2}\in \mathbb{R}^{4 \times W \times H}$, $\mbf{s}_{3}\in \mathbb{R}^{6 \times W \times H}$, $\mbf{s}_{4}\in \mathbb{R}^{13 \times W \times H}$.
These score maps have two properties: (a) they can serve as a language prior, which enables us to leverage the powerful image-text association ability of the pre-trained CLIP model; (b) being supervised against the corresponding ground truth manipulation mask, these manipulation score maps learn to indicate the region that is manipulated by the given forgery method, offering auxiliary signals for manipulation localization.

These generated 2D score maps are incorporated in the existing HiFi-Net and help localize the manipulation area, as depicted in Fig.~\ref{fig:architecture}. Specifically, we first concatenate the HiFi-Net output features (\textit{e.g.}, $\net_b$), up-sampling visual embeddings $\mbf{V}_{b\uparrow}$---where the arrow indicates upsampling operation---and clip image encoder feature $\mathbf{Z}$ into a final fused feature map $\mbf{Y}_b$. 
\begin{equation}
  \mbf{Y}_b \doteq [\mathbf{Z}_{\uparrow},\mathbf{V}_{b\uparrow},\net_b] \quad  b \in \{1 \ldots 4\}.  
\end{equation}

\Paragraph{Discussion} It is important to alleviate the potential concern about why the pre-trained CLIP image and text encoders generate manipulation score maps that help the forgery localization tasks, as the pre-trained CLIP model is meant to correlate image and text based on their semantics. In fact, the manipulation score map is generated by $\mathbf{F}_{con}$ and $\mathbf{T}^{\prime}_{b}$ via Eq.~(\ref{eq_score_map}), and this $\mathbf{F}_{con}$ is the concatenated feature of $\mathbf{V}_{b}$, which are empirically generalize to the manipulation localization task (details in Tab.~\ref{table_tax_loc} of Sec.~\ref{ex_data_IFDL}). In addition, text embedding itself contains rich semantics and we further use a refinement module to make it better adapt for our manipulation localization task. Note that we define a pixel-wise loss (\textit{e.g.}, cross entropy) between the manipulation score map and ground truth forgery mask. We use this pixel-wise loss to optimize the trainable refinement module.

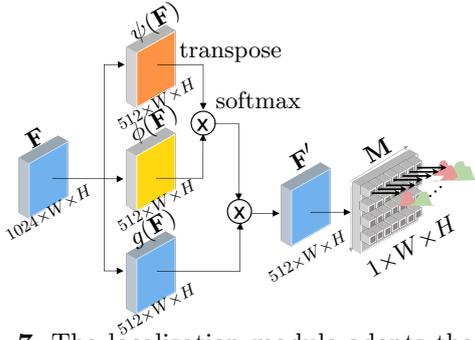
\begin{figure}[t]
    \centering
    \begin{overpic}[width=0.4\textwidth]{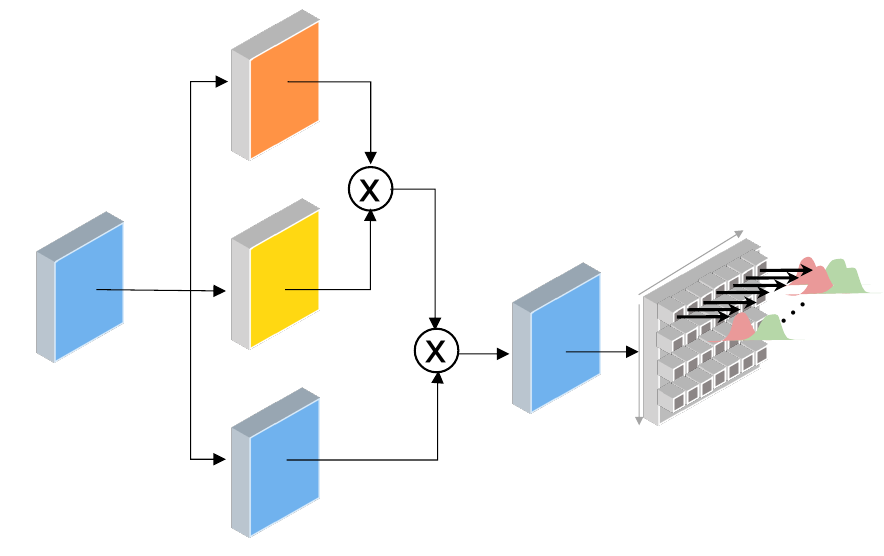}
    \put(5,36){\rotatebox{30}{\small{$\mathbf{F}$}}}
    \put(1,17){\scalebox{.7}{\rotatebox{30}{\small{$1024\PLH W \PLH H$}}}}
    
    \put(26,58.5){\rotatebox{30}{\scriptsize{$\psi(\mathbf{F})$}}}
    \put(24,40){\scalebox{.7}{\rotatebox{30}{\small{$512\PLH W \PLH H$}}}}
    \put(26,37.5){\rotatebox{30}{\scriptsize{$\mathbf{\phi(F)}$}}}
    \put(24,19){\scalebox{.7}{\rotatebox{30}{\small{$512\PLH W \PLH H$}}}}
    \put(26,16){\rotatebox{30}{\scriptsize{$g(\mathbf{F})$}}}
    \put(24,-2.5){\scalebox{.7}{\rotatebox{30}{\small{$512\PLH W \PLH H$}}}}
    \put(37,55){\scriptsize{transpose}}
    \put(44.5,44.5){\scriptsize{softmax}}
    \put(59,31.5){\rotatebox{30}{\small{$\mathbf{F^\prime}$}}}
    \put(55,8.5){\scalebox{.7}{\rotatebox{30}{\small{$512\PLH W \PLH H$}}}}
    \put(74.5,33){\rotatebox{30}{\small{$\mathbf{M}$}}}
    \put(75,9.5){\scalebox{1}{\rotatebox{30}{\scriptsize{$1\PLH W \PLH H$}}}}
    \end{overpic}
    \caption{The localization module adopts the self-attention mechanism to transfer the feature map $\mathbf{F}$ to the localization mask $\mathbf{M}$. 
    }\label{fig_localization_module}
\end{figure}

\SubSection{Localization Module}
\label{sec_loc}
\Paragraph{Architecture} 
We first feed the final fused $\mathbf{Y}_{b}$ into the feature pyramid network~\cite{lin2017feature}, which is widely used as the effective algorithm for fusing multi-scale feature maps in the pixel-wise prediction task. After that, we have the final fused feature map, denoted as $\mathbf{F}\in \mathbb{R}^{512\PLH W \PLH H}$, which is used to output the mask $\hat{\mbf{M}}$ for localizing the forgery.

To model the dependency and interactions of pixels on the large spatial area, the localization module employs the self-attention mechanism~\cite{zhang2019self_att_gan,wang2018non_local}. 
As shown in the localization module architecture in Fig.~\ref{fig_localization_module}, 
we use $1\PLH 1$ convolution to form $g$, $\phi$ and $\psi$, which convert input feature $\mathbf{F}$ into $\mathbf{F}_{g} = g(\mathbf{\mathbf{F}})$, $\mathbf{F}_{\phi} = \phi(\mathbf{\mathbf{F}})$ and $\mathbf{F}_{\psi} = \psi(\mathbf{\mathbf{F}})$. Given $\mathbf{F}_{\phi}$ and $\mathbf{F}_{\theta}$, we compute the spatial attention matrix $\mathbf{A}_{s} = \operatorname{softmax} (\mathbf{F}_{\phi}^{T} \mathbf{F}_{\theta})$. We then use this transformation $\mathbf{A}_{s}$ to map $\mathbf{F}_{g}$ into a global feature map  $\mbf{F}^{\prime}=\mathbf{A}_{s} \mathbf{F}_{g} \in \mathbb{R}^{512\PLH W \PLH H}$.

\Paragraph{Objective Function} Following~\cite{masi2020two}, we employ a metric learning objective function for localization, which creates a wider margin between real and manipulated pixels.
We first learn features of each pixel and then model the geometry of such learned features with a radial decision boundary in the hyper-sphere. Specifically, we start with pre-computing a reference center $\mathbf{c} \in \mathbb{R}^{D}$, by averaging the features of all pixels in real images of the training set.
We use $\mbf{F}^{\prime}_{ij}\in \mathbb{R}^{D}$ to indicate the $ij$-th pixel of the final mask prediction layer. Therefore, our localization loss $\mathcal{L}_{loc}$ is:
\begin{equation}
\small
\mathcal{L}_{loc} = \frac{1}{HW} \sum_i^H\sum_j^W \mathcal{L}\big(\mbf{F}^{\prime}_{ij}, \mbf{M}_{ij}; \mathbf{c}, \tau\big),
\vspace{-3mm}
\label{eq_L_loc}
\end{equation}
where:
\begin{equation*}
\small
\mathcal{L} = \begin{cases} \left \| \mbf{F}^{\prime}_{ij} - \mathbf{c} \right \|_2 & \mbox{if } 
 \mbf{M}_{ij} \mbox{ real} \\ 
 \max\big(0, \tau - \left \| \mbf{F}^{\prime}_{ij} - \mathbf{c} \right \|_2\big) & \mbox{if } \mbf{M}_{ij}\mbox{ forged}. \end{cases}
\label{eq_deep_loss}
\end{equation*}
Here $\tau$ is a pre-defined margin. 
The first term in $\mathcal{L}$ improves the feature space compactness of real pixels. 
The second term encourages the distribution of forged pixels to be far away from real by a margin $\tau$. 
Note our method differs to \cite{ruff2018deep,masi2020two} in two aspects: 1) unlike~\cite{ruff2018deep}, we use the second term in $\mathcal{L}$ to enforce separation; 2) compared to the image-level loss in~\cite{masi2020two} that has two margins, we work on the more challenging pixel-level learning. Thus, we use a single margin, which reduces the number of hyper-parameters and improves the simplicity.

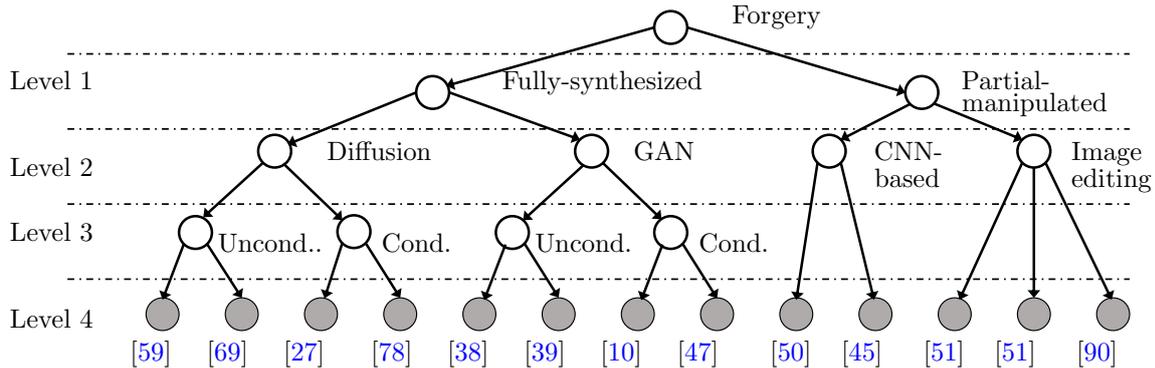
\begin{figure*}[t]
    \centering
    \begin{overpic}[width=0.9\linewidth]{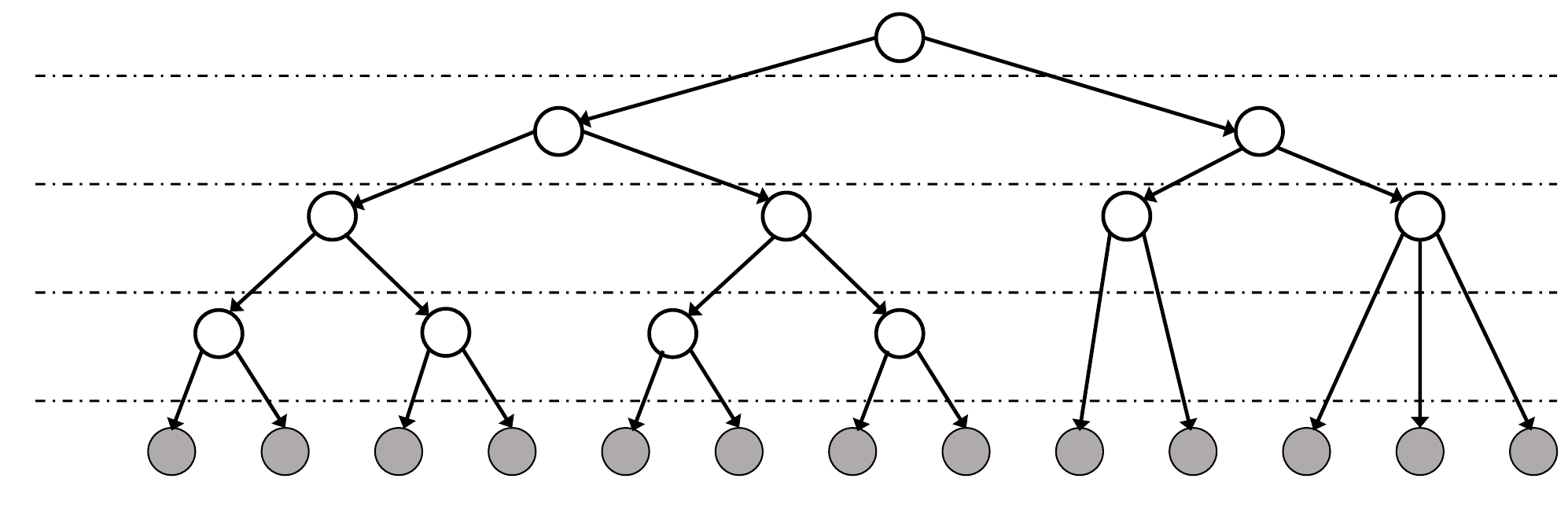}
    \put(-3,25){{Level $1$}}
    \put(-3,17){{Level $2$}}
    \put(-3,11){{Level $3$}}
    \put(-3,3){{Level $4$}}
    
    \put(63,31){{Forgery}}
    \put(42,25){{Fully-synthesized}}
    \put(84,25){{Partial-}}
    \put(84,23){{manipulated}}
    
    \put(26,18.5){{Diffusion}}
    \put(54,18.5){{GAN}}
    \put(76,18.5){{CNN-}}
    \put(76,16){{based}}
    \put(94,18.5){{Image}}
    \put(94,16){{editing}}
    
    \put(16,10){{Uncond..}}
    \put(31,10){{Cond.}}
    \put(45,10){{Uncond.}}
    \put(60,10){{Cond.}}
    
    \put(8,0){{\cite{nichol2021glide}}}
    \put(15,0){{\cite{rombach2021highresolution_latent_diffusion}}}
    \put(22,0){{\cite{ho2020denoising_ddpm}}}
    \put(30,0){{\cite{song2020denoising_ddim}}}
    
    \put(37,0){{\cite{karras2020training_styleGan2ADA}}}
    \put(44,0){{\cite{karras2021alias_styleGan3}}}
    \put(51,0){{\cite{choi2020starganv2}}}
    \put(58,0){{\cite{li2021image_hisd}}}
    
    \put(66.5,0){{\cite{liu2019stgan}}}
    \put(73,0){{\cite{li2019faceshifter}}}
    
    \put(80.5,0){{\cite{liu2022pscc}}}
    \put(87,0){{\cite{liu2022pscc}}}
    \put(94.5,0){{\cite{wu2018busternet_copy_move_yue_wu}}}
    \end{overpic}
    \vspace{1mm}
    \caption{Overview of HiFi-IFDL dataset. At level $1$, we separate forged images into fully-synthesized and partial-manipulated. At level $2$, we discriminate different forgery methodologies, \textit{e.g.}, image editing, CNN-based partial manipulation, Diffusion or GANs. 
    Then, at level $3$, we separate images based on whether forgery methods are conditional or unconditional. The final level $4$ refers to the specific forgery method. [Key: Uncond.: unconditional, Cond.: conditional].}
    \label{fig_taxonomy_benchmark}
\end{figure*}

\SubSection{Training and Inference}\label{sec_train_infer}
In the training, each branch is optimized towards the classification at the corresponding level, we use $4$ classification losses, $\mathcal{L}^{1}_{cls}$, $\mathcal{L}^{2}_{cls}$, $\mathcal{L}^{3}_{cls}$ and $\mathcal{L}^{4}_{cls}$ for $4$ branches. At the branch $b$, $\mathcal{L}^{b}_{cls}$ is the cross entropy distance between $p(\mbf{y}_b|\mbf{X})$ and a ground truth categorical $\hat{\mbf{y}}_b$.
When the input image is labeled as ``real'', we only apply the last branch ($\net_{4}$) loss function. Otherwise, we use all the branches. Let us denote the input image as $\mbf{X}$.
\begin{equation*}
\mathcal{L}_{cls} = 
    \begin{cases} 
        \mathcal{L}^{1}_{cls} + \mathcal{L}^{2}_{cls} + \mathcal{L}^{3}_{cls} + \mathcal{L}^{4}_{cls} & \mbox{if $\mbf{X}$ is forged}\\ 
        \mathcal{L}^{4}_{cls} & \mbox{if $\mbf{X}$ is real.}
    \end{cases}
\end{equation*}
In addition, we compute the cross entropy between the predicted manipulation score map $\mathbf{s}_{b}$ and the downsampled manipulation label $\mathbf{M}_{\downarrow}$. We denote this loss as $\mathcal{L}^{score}_{loc}$. The architecture is trained end-to-end with different learning rates per layer. The detailed objective function is:
\begin{equation}
    \mathcal{L}_{tot} = \lambda_{1} \mathcal{L}_{loc} + \lambda_{2} \mathcal{L}_{cls} + \lambda_{3} \mathcal{L}^{score}_{loc}. 
    \label{eq:objective_function}
\end{equation}
$\lambda_{1}$, $\lambda_{2}$, and $\lambda_{3}$ are hyper-parameters that keep different objective terms on the reasonable magnitudes.

In the inference, HiFi-Net++ takes the input image $\mathbf{X}$ and pre-defined text input $\mathbf{T}_{b}$, and then generate the forgery mask from the localization module, and predicts forgery attributes at different levels. 
We use the output probabilities at level $4$ for forgery attribute classification. 
For binary ``forged vs.~real'' classification, we predict as forged if the highest probability falls in any manipulation method at level $4$.


\begin{figure*}[t]
    \centering
    \begin{subtable}[b]{0.55\linewidth}
        \centering
        \resizebox{0.98\textwidth}{!}{
            \small 
            \begin{tabular}{c|c|c|c}\hline
                Forgery Method&Image Source&Images $\#$&Source\\ \hline\hline
                DDPM~\cite{ho2020denoising_ddpm}& LSUN& $100$k & pre-trained\\ \hline
                DDIM~\cite{song2020denoising_ddim}& LSUN& $100$k & pre-trained\\ \hline
                GDM.~\cite{nichol2021glide}& LSUN& $100$k & pre-trained\\ \hline
                LDM.~\cite{rombach2021highresolution_latent_diffusion}& LSUN& $100$k & pre-trained\\ \hline
                StarGANv$2$~\cite{choi2020starganv2}& CelebaHQ& $100$k & pre-trained\\ \hline
                HiSD~\cite{li2021image_hisd}& CelebaHQ& $100$k & pre-trained\\ \hline
                StGANv$2$-ada~\cite{karras2020training_styleGan2ADA}& FFHQ, AFHQ& $100$k & pre-trained\\ \hline
                StGAN$3$~\cite{karras2021alias_styleGan3}& FFHQ, AFHQ& $100$k & pre-trained\\ \hline
                STGAN~\cite{liu2019stgan}& CelebaHQ& $100$k & self-train\\ \hline
                Faceshifter~\cite{li2019faceshifter}& Youtube video& $100$k & released\\ \hline
                \hline
                \end{tabular}}
        \caption{}
        \label{tab_dataset}
    \end{subtable}
    \begin{subfigure}[b]{0.4\linewidth}
    \centering
        \includegraphics[width=0.95\linewidth]{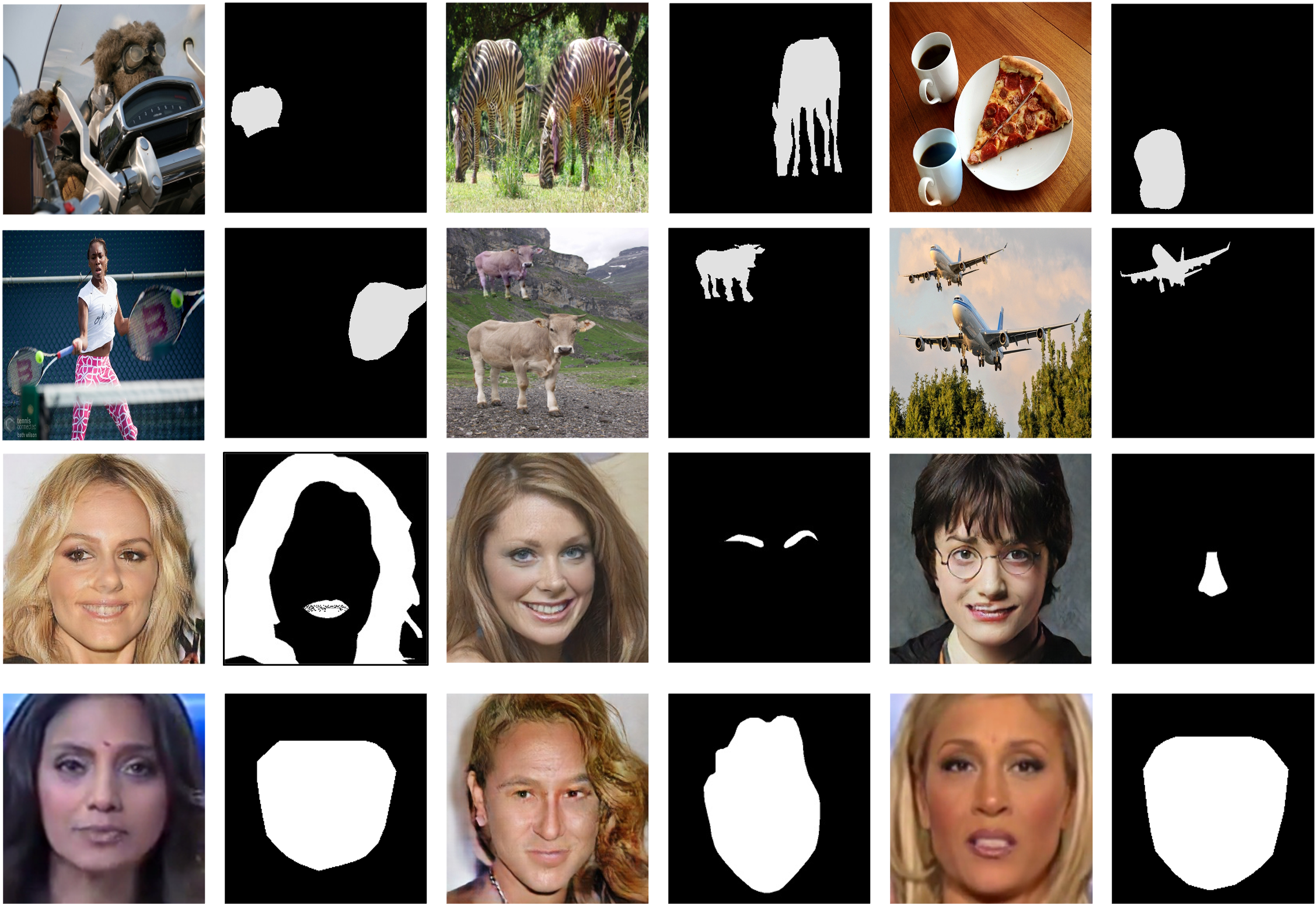}
        \caption{}
        \label{fig_taxonomy_mask}
    \end{subfigure}
    \caption{The details of the collected dataset. Each column in order shows forgery method; the image source used for the generation; the image number; if images are generated with pre-trained/self-trained models or released images.}
\end{figure*}

\Section{Hierarchical Fine-grained IFDL dataset}\label{sec_benchmark}
We construct a fine-grained hierarchical benchmark, named HiFi-IFDL, to facilitate our study. 
HiFi-IFDL contains some most updated and representative forgery methods, for two reasons: 1) Image synthesis evolved into a more advanced era and artifacts become less prominent in the recent forgery method; 2) It is impossible to include all possible generative method categories, such as VAE~\cite{kingma2013auto_vae_gen} and face morphing~\cite{scherhag2019face_morphing}. 
So we only collect the most-studied forgery types (\textit{i.e.,} splicing) and the recent generative methods (\textit{i.e.,} DDPM).

Specifically, HiFi-IFDL includes images generated from $13$ forgery methods spanning from CNN-based manipulations to image editing, and these forged methods are based on taxonomy illustrated in Fig.~\ref{fig_taxonomy_benchmark}.
More formally, we use Tab.~\ref{tab_dataset} to report more details of these forgery methods, each of which generates $100,000$ images.  
For the real images,  we select them from $6$ datasets (\textit{e.g.,} FFHQ~\cite{karras2019style}, AFHQ~\cite{choi2020starganv2}, CelebaHQ~\cite{lee2020maskgan_celebahq}, Youtube face~\cite{rossler2019faceforensics++v3}, MSCOCO~\cite{lin2014microsoft_coco}, and LSUN~\cite{yu2015lsun}). 
More formally, for datasets (\textit{e.g.}, LSUN, and YouTube videos) that contain more than $100,000$ images or video frames, we randomly select $100,000$ images from them. For datasets (\textit{e.g.}, FFHQ~\cite{karras2019style}, AFHQ~\cite{choi2020starganv2}, and CelebaHQ~\cite{lee2020maskgan_celebahq}) with a total number of images less than $100,000$, we utilize the entire dataset to construct the HiFi-IFDL.
As a result, training, validation, and test sets have $1,710$K, $15$K, and $174$K images, respectively. 
Constructing such a large-scale dataset is essential, as it offers a wide variety of forgery patterns and realistic distributions of real images, which help train a robust detection model that performs effectively in real-world scenarios.
In Fig.~\ref{fig_taxonomy_mask}, we show several examples taken from our dataset that represent a variety of objects, faces, and animals. All manipulations are marked by the high-resolution binary mask. 

While there are different ways to design a forgery hierarchy, our hierarchy starts at the root of an image being forged, and then each level is made more and more specific to arrive at the actual generator. 
Our work studies \emph{the impact of the hierarchical formulation to IFDL}. While different hierarchy definitions are possible, it is beyond the scope of this paper.

\Section{Experiments}\label{sec_exp}
In this section, we first evaluate IFDL performance of the proposed HiFi-Net++ on HiFi-IFDL dataset and report the results in Sec.~\ref{ex_data_IFDL}. Then, in Sec.~\ref{ex_data_editing}, we examine the IFDL performance on the image editing domain, following the experiment setup defined in~\cite{wang2022objectformer}, and compare to various prior works~\cite{wu2019mantra,hu2020span,liu2022pscc,dong2022mvss,wang2022objectformer,guo2023hierarchical} on $5$ different datasets, including \textit{Columbia}~\cite{ng2009columbia}, \textit{Coverage}~\cite{wen2016coverage}, \textit{CASIA}~\cite{dong2013casia} and \textit{NIST$\textit{16}$}~\cite{NIST16}. 
After that, in Sec.~\ref{ex_cnn_det}, we conduct a comparison to SoTA forgery detection methods on identifying images generated by various forgery methods.
In the end, for the digital facial manipulated image, we conduct the evaluation on the Diverse Fake Face Dataset (DFFD) dataset~\cite{stehouwer2019detection}, comparing it with prior works~\cite{stehouwer2019detection}. The reason we choose DFFD is because DFFD offers the manipulation mask label on the fake facial images, which enables us to measure both detection and localization performance. 

\subsection{Implementation Details}
HiFi-Net++ is implemented on PyTorch and, during the training procedure, trained with $40$ epochs and each epoch contains $100,000$ iterations. The training batch size is $16$, with $8$ real and $8$ forged images.
In our HiFi-Net++, multi-branch feature extractor, the feature map resolutions for different branches are $256$, $128$, $64$, and $32$ pixels. In the experiment on the HiFi-IFDL dataset, the fine-grained classification for $1$st, $2$nd, $3$rd, and $4$th levels are $2$-way, $4$-way, $6$-way, and $14$-way multi-class classification, respectively. In terms of the Language-guided Forgery Localization Enhancer, we have experimented with different image encoder backbones, such as ResNet-$50$, ResNet-$101$ and ViT-B-$16$. The overall performance from these backbones does not vary largely, and ResNet$50$ offers the best computation efficiency. Therefore, we select ResNet-$50$ as the pre-trained CLIP image encoder for the main result analyzed in the rest of the paper, except the ablation study.

As for the details of $\mathcal{L}_{loc}$ implementation, we first use the Feature Pyramid Network to convert the joint feature $\mathbf{Y}_{b}$ to the high-dimensional feature $\mathbf{F}^{\prime}_{ij} \in R^{D}$, where $D = 64$. Then we average the feature $\mathbf{F}^{\prime}_{ij} \in R^{D}$ for all pixels in the real image from the HiFi-IFDL, and this average value then is used as $\mathbf{c}$. Then, we compute the $\ell_{2}$ distance between each pixel feature $\mathbf{F}^{\prime}_{ij} \in R^{D}$ and $\mathbf{c}$, and denote the largest distance as $D_{max}$. In the Eq.~\ref{eq_L_loc} of the paper, we set the threshold $\tau$ as $2.5\cdot D_{max}$.

Both HiFi-Net~\cite{guo2023hierarchical} and HiFi-Net++ are trained in an end-to-end manner. The detailed hyper-parameters in Eq.~\ref{eq:objective_function} are $\lambda_1 = 10$, $\lambda_2 = 1$, and $\lambda_3 = 0.01$. The feature extractor is modified based on the pre-trained HRNet~\cite{wang2020deepHRNet}, in which we add color and frequency-enhanced blocks, as well as more convolution layers. Consequently, each branch of our multi-branch feature extractor can have an identical number of convolutional layers, and such details can be found in our source code, which is publicly released. Moreover, we employ different learning rates for different trainable modules. For example, we use $\texttt{1e-4}$ for the multi-branch feature extractor and classification module, $\texttt{3e-4}$ for the localization module, and $\texttt{5e-5}$ for the refinement module inside the language-guided forgery localization enhancer. Note that we do not change the learning strategy when different pre-trained CLIP image encoders are used in the language-guided forgery localization enhancer. During the training, we use $\texttt{ReduceLROnPlateau}$ as the learning rate scheduler to reduce the learning rate when the value of the composite loss function on the validation set does not decrease for continuous $75$ times.

\begin{table}[t]
\begin{subtable}{1.\linewidth}
    \centering
        \resizebox{1.\textwidth}{!}{
        \begin{tabular}{c|cccc|cc}
        \hline
        \multirow{2}{*}{\begin{tabular}{c}
            \textbf{ Forgery}\\
            \textbf{ Detection}
        \end{tabular}}&\multicolumn{2}{c}{CNN-syn.}&\multicolumn{2}{c|}{Image Edit.}&\multicolumn{2}{c}{Overall}\\\cline{2-7}
        &AUC &F$1$ &AUC &F$1$ &AUC &F$1$\\ \hline
        CNN-det.$^*$~\cite{wang2020cnn}&$76.5$&$60.5$&$54.8$&$33.5$&$56.5$&$40.5$\\
        CNN-det.~\cite{wang2020cnn}&$92.3$&$90.0$&$87.0$&$74.7$&$90.1$&$83.7$\\
        Two-bran.~\cite{masi2020two}&$93.3$&$89.2$&$83.3$&$66.7$&$86.7$&$80.2$\\
        Att. Xce.~\cite{stehouwer2019detection}&$93.8$&$91.2$&$90.8$&$82.1$&$87.3$&$90.0$\\
        PSCC-Net~\cite{liu2022pscc}&$94.6$&$93.2$&$90.7$&$82.3$&$93.2$&$91.3$\\
        Uni-Det~\cite{ojha2023towards}&$94.0$&$94.2$&$63.2$&$41.5$&$77.2$&$60.2$\\
        \hline\hline

        HiFi-Net~\cite{guo2023hierarchical}&$97.0$&$\mathbf{96.1}$&$91.5$&$85.9$&$96.8$&$\mathbf{94.1}$\\
        HiFi-Net++&$\mathbf{97.8}$&$95.0$&$\mathbf{91.7}$&$\mathbf{86.2}$&$\mathbf{97.2}$&$93.7$\\ \hline

        \end{tabular}
        }
    \caption{CNN-detector~\cite{wang2020cnn} has $4$ variants with different augmentations, and we report the variant with the best performance. For Two-branch~\cite{masi2020two}, we implement this method with the help of its authors.}
    \label{table_tax_det}
\end{subtable}
\begin{subtable}{1.\linewidth}
    \centering
    \resizebox{1.\textwidth}{!}{
    \begin{tabular}{c|cccc|cc}
    \hline
    \multirow{2}{*}{\begin{tabular}{c}
          \textbf{ Forgery}\\
        \textbf{ Localization}
    \end{tabular}}&\multicolumn{2}{c}{CNN-syn.}&\multicolumn{2}{c|}{Image Edit.}&\multicolumn{2}{c}{Overall}\\\cline{2-7}
    &AUC &F$1$ &AUC &F$1$ &AUC &F$1$\\ \hline
    OSN-det.$^*$~\cite{wu2022robust}&$51.4$&$38.8$&$83.2$&$70.1$&$79.4$&$56.5$\\    CatNet$^*$~\cite{kwon2022learning}&$48.6$&$31.9$&$86.1$&$79.4$&$78.3$&$65.1$\\
    CatNet~\cite{kwon2022learning}&$92.5$&$81.5$&$92.0$&$88.2$&$92.4$&$86.8$\\
    Att. Xce.~\cite{stehouwer2019detection}&$89.1$&$87.7$&$83.3$&$79.3$&$87.1$&$86.5$\\
    PSCC-Net~\cite{liu2022pscc}&$94.3$&$96.8$&$91.1$&$86.5$&$92.7$&$94.9$\\
    \hline
    \hline
    HiFi-Net~\cite{guo2023hierarchical}&$98.4$&$97.0$&$93.0$&$90.1$&$95.3$&$96.9$\\ \hline
    CLIP-ResNet$50$$^{1}$~\cite{radford2021learning}&$90.2$&$80.6$&$84.5$&$79.3$&$88.7.3$&$84.2$\\ 
    CLIP-ViT-B-16$^{1}$~\cite{radford2021learning}&$87.5$&$78.3$&$79.8$&$73.2$&$80.2$&$69.7$\\ \hline
    HiFi-Net++&$\mathbf{98.6}$&$\mathbf{98.1}$&$\mathbf{97.8}$&$\mathbf{95.7}$&$\mathbf{97.7}$&$\mathbf{98.6}$\\ 
    \hline
    \end{tabular}
    }
    \caption{OSN-det~\cite{wu2022robust} only releases pre-trained weights with the inference script, without the training script. CLIP image encoders with $^{1}$ represents that we freeze pre-trained CLIP image encoders and only train a few transposed convolution layers.}
    \label{table_tax_loc}
\end{subtable}
\begin{subfigure}{1\linewidth}
    \centering
    \includegraphics[scale=0.28]{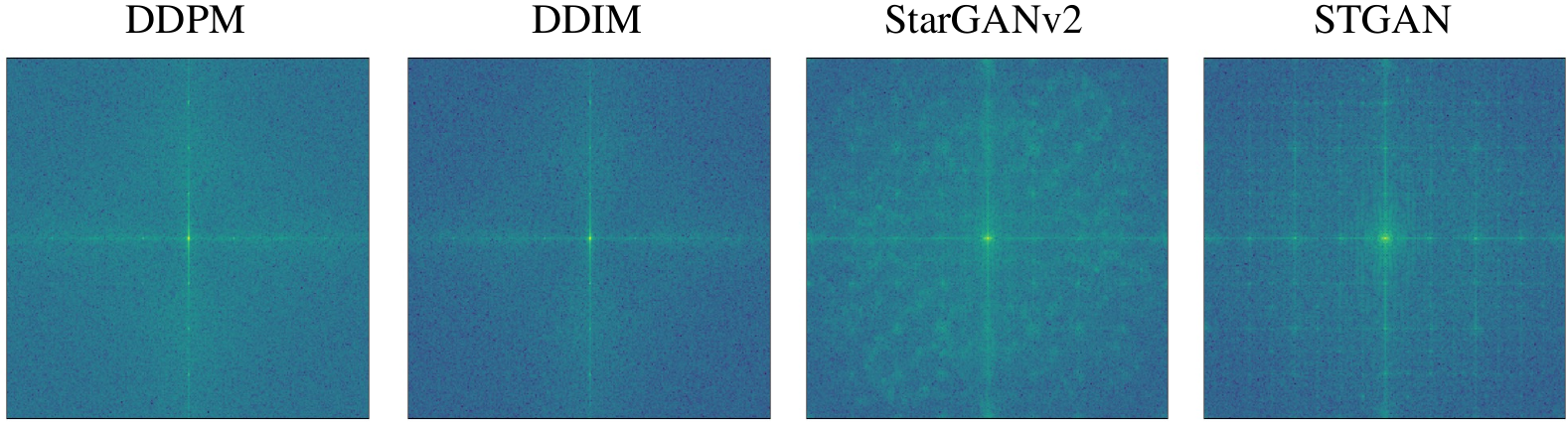}
    \caption{Frequency artifacts in different forgery methods. DDPM~\cite{ho2020denoising_ddpm} and DDIM~\cite{song2020denoising_ddim} do not exhbit the checkboard patterns~\cite{zhangxue2019detecting,wang2020cnn} observed in GAN-based methods, such as StarGAN-v$2$~\cite{choi2020starganv2} and STGAN~\cite{liu2019stgan}.}
    \label{table_tax_artifacts}
\end{subfigure}
\caption{IFDL Results on HiFi-IFDL. $*$ means we apply author-released pre-trained models. Models without $*$ mean they are trained on HiFi-IFDL training set. [\textbf{Bold}: best result].}
\label{table_tax_IMDL}
\end{table}

\begin{figure}[t]
    \centering
    \includegraphics[width=0.46\textwidth]{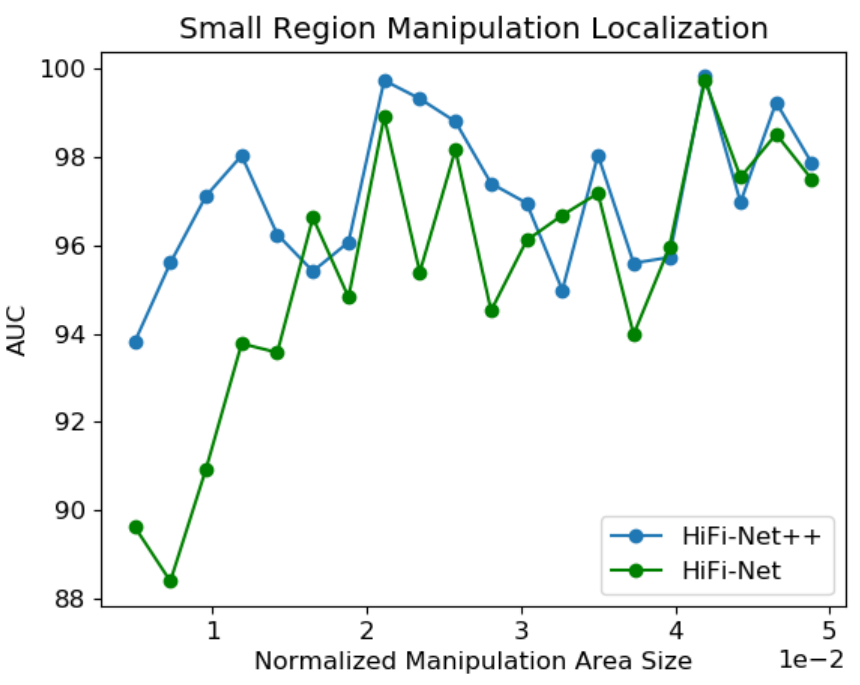}
    \caption{The small region manipulation localization performance of HiFi-Net~\cite{guo2023hierarchical} and HiFi-Net++. The X-axis denotes the normalized manipulation area size, indicating this plot only visualizes forged images with manipulation less than $5\%$ of the input image size. The Y-axis depicts the average AUC of two methods achieve among images with the given manipulation area size. With the help of the language-guided forgery localization enhancer, the HiFi-Net++ improves the localization performance over HiFi-Net, on images with small manipulation regions.}
    \label{fig:small_region}
\end{figure}
\begin{figure*}[t]
    \centering
    \includegraphics[width=1.\linewidth]{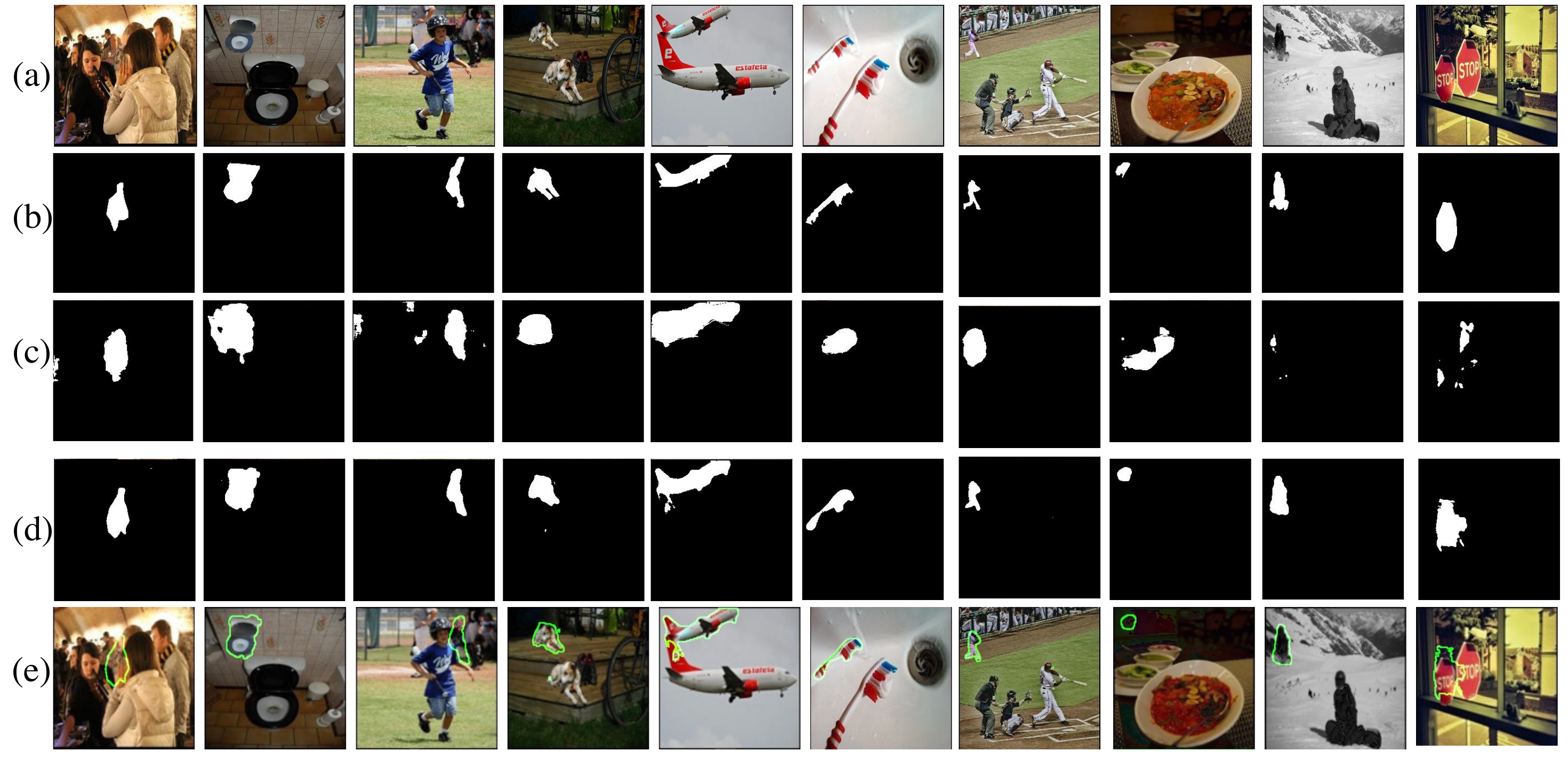}
    \caption{The qualitative localization performance on the image editing domain data in HiFi-IFDL \textit{test} dataset. From top to down, rows are (a) input image, (b) ground truth mask, (c) pre-trained CLIP image encoder (ResNet-$50$) localization result, (d) HiFi-Net++ localization result, and (e) Overlaid image based on the HiFi-Net++ localization result. From row (c), the pre-trained CLIP image encoder can roughly locate the manipulation objects, whereas it still sometimes identifies the manipulation in the wrong way (last three images in this row). In contrast, in row (d) and (e), we show the HiFi-Net++ can produce localization mask that is close to the ground truth and identifies the manipulation region accurately.}
    \label{fig_viz_hifi_editing}
\end{figure*}

\SubSection{HiFi-IFDL Dataset Performance}\label{ex_data_IFDL}

Tab.~\ref{table_tax_IMDL} reports the different model performance on the HiFi-IFDL dataset, in which we use AUC and F$1$ score as metrics on both image-level forgery detection and pixel-level localization. Specifically, in Tab.~\ref{table_tax_det}, first, we observe that the pre-trained CNN-detector~\cite{wang2020cnn} does not perform well because it is trained on GAN-generated images that are different from images manipulated by diffusion models. Such differences can be seen in Fig.~\ref{table_tax_artifacts}, where we visualize the frequency domain artifacts by following the routine~\cite{wang2020cnn} that applies the high-pass filter on the image generated by different forgery methods. 
Similar visualization is adopted in~\cite{wang2020cnn,zhangxue2019detecting,corvi2022detection,ricker2022towards} also.
Then, we train both prior methods on HiFi-IFDL, and they again perform worse than our model: CNN-detector uses ordinary ResNet$50$, but our model is specifically designed for image forensics. 
Two-branch processes deepfakes video by LSTM that is less effective in detecting forgery in the image editing domain. Attention Xception~\cite{stehouwer2019detection} and PSCC-Net are proposed for facial image forgery and image editing domains, respectively. 
These two methods perform worse than HiFi-Net++ by $9.9\%$ and $4.0\%$ AUC, respectively. In the last two rows, we can observe that HiFi-Net++ achieves $0.4\%$ higher AUC and $0.4\%$ lower F$1$ scores than the HiFi-Net. It is reasonable for HiFi-Net++ to have comparable image forgery detection performances with HiFi-Net, since the newly-proposed language-guided forgery localization enhancer mainly improves the localization performance, and features used for forgery image detection are the same in the HiFi-Net and HiFi-Net++.

In Tab.~\ref{table_tax_loc}, we compare with previous methods which can perform the forgery localization. Specifically, the pre-trained OSN-detector~\cite{wu2022robust} and CatNet~\cite{kwon2022learning} do not work well on CNN-synthesized images in HiFi-IFDL dataset, since they merely train models on images manipulated by editing methods. 
Then, we use HiFi-IFDL dataset to train CatNet, but it still performs worse than ours: CatNet uses DCT stream to help localize areas of splicing and copy-move, but HiFi-IFDL contains more forgery types (\textit{e.g.}, inpainting). Meanwhile, the accurate classification performance further helps the localization as statistics and patterns of forgery regions are related to different individual forgery methods. For example, for the forgery localization, the HiFi-Net achieves $2.6\%$ AUC and $2.0\%$ F$1$ improvement over PSCC-Net. Additionally, the superior localization demonstrates that our hierarchical fine-grained formulation learns more comprehensive forgery localization features than the multi-level localization scheme proposed in PSCC-Net. 
\begin{figure*}[ht]
    \centering
    \includegraphics[width=1.\linewidth]{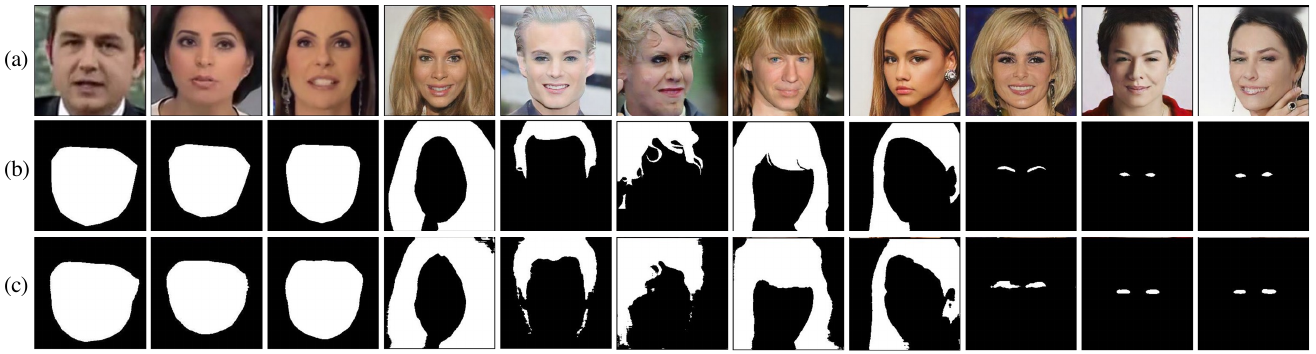}
    \caption{The qualitative manipulation localization performance on the facial forgery images in HiFi-IFDL \textit{test} dataset. From top to bottom, figures are (a) input image, (b) ground truth localization mask, and (c) localization result of HiFi-Net++. As depicted in the last row, HiFi-Net++ is able to identify manipulation on different facial components, such as the face and hair of different styles. Also, the small manipulations on eyebrows (the third last column) and eyes (last two columns) can be identified.}
    \label{fig_viz_hifi_face}
\end{figure*}

Before discussing the HiFi-Net++, we first report how the pre-trained CLIP encoder generalizes to the manipulation localization task. First, we obtain the intermediate features $\mathbf{V}_{b} \in \mathbb{R}^{C \times W_{b} \times H_{b}} \text{ with } b \in \{1 \ldots 4\}$ from the pre-trained CLIP image encoder (Sec.~\ref{sec_lang_guided_fe} and Fig.~\ref{fig:architecture}). 
We apply a $1\PLH 1$ convolution (denoted as $\mathbf{w}_{1\times1}$) to convert the channel-wise dimension $C$ of each intermediate feature into $64$, and upsample $\mathbf{V}_{4}$, $\mathbf{V}_{3}$, and $\mathbf{V}_{2}$, such that they have the same width and height as $\mathbf{V}_{1}$. Then, we concatenate these transformed feature maps into a joint visual embedding, on which we apply a few transposed convolution layers for localizing the manipulation. During the training, we only optimize a few transposed convolution layers and $\mathbf{w}_{1\times1}$, while the pre-trained CLIP image encoder remains fixed. Consequently, as shown in the Tab.~\ref{table_tax_loc}, CLIP-ResNet$50$ achieves similar localization results as Attention Xception~\cite{stehouwer2019detection}, which is exclusively trained on the HiFi-IFDL dataset. Furthermore, we show the qualitative results of its localization performance in Fig.~\ref{fig_viz_hifi_editing}, in which pre-trained CLIP image encoder can capture the rough shape and only fails to capture the specific contour of objects, (\textit{e.g.}, person). This demonstrates that the pre-trained CLIP image encoder has the generalization ability to help the manipulation localization. Although CLIP-ViT-B-$16$ does not offer as impressive results as CLIP-ResNet-$50$, it still yields comparable results with the pre-trained CatNet which specializes in localizing manipulations.

Therefore, with the additional language-guided forgery localization module, which contains the pre-trained CLIP image encoder, HiFi-Net++ achieves the better overall localization performance over the HiFi-Net, and this improvement mainly comes from the image editing domain where HiFi-Net++ achieves $4.8\%$ higher AUC and $5.6\%$ F$1$ score over HiFi-Net. We believe this is because, in the language-guided forgery localization module, both the pre-trained CLIP visual embedding and manipulation score maps play key roles in this improvement. More details can be found in the ablation study (Sec.~\ref{ex_ablation}).
Also, even though HiFi-Net++ does not largely surpass over the HiFi-Net on the localization performance of the CNN-synthesized image, it again achieves the better performance in both AUC and F$1$ scores. Moreover, HiFi-Net++ delivers the better performance on localizing the small region manipulations, as illustrated in Fig.~\ref{fig:small_region}. Qualitatively, we showcase the manipulation localization results of HiFi-Net++ in Fig.~\ref{fig_viz_hifi_editing} and Fig.~\ref{fig_viz_hifi_face}. Based on these empirical results, we can conclude that, by introducing the additional language-guided forgery localization enhancer, the localization performance can be enhanced.

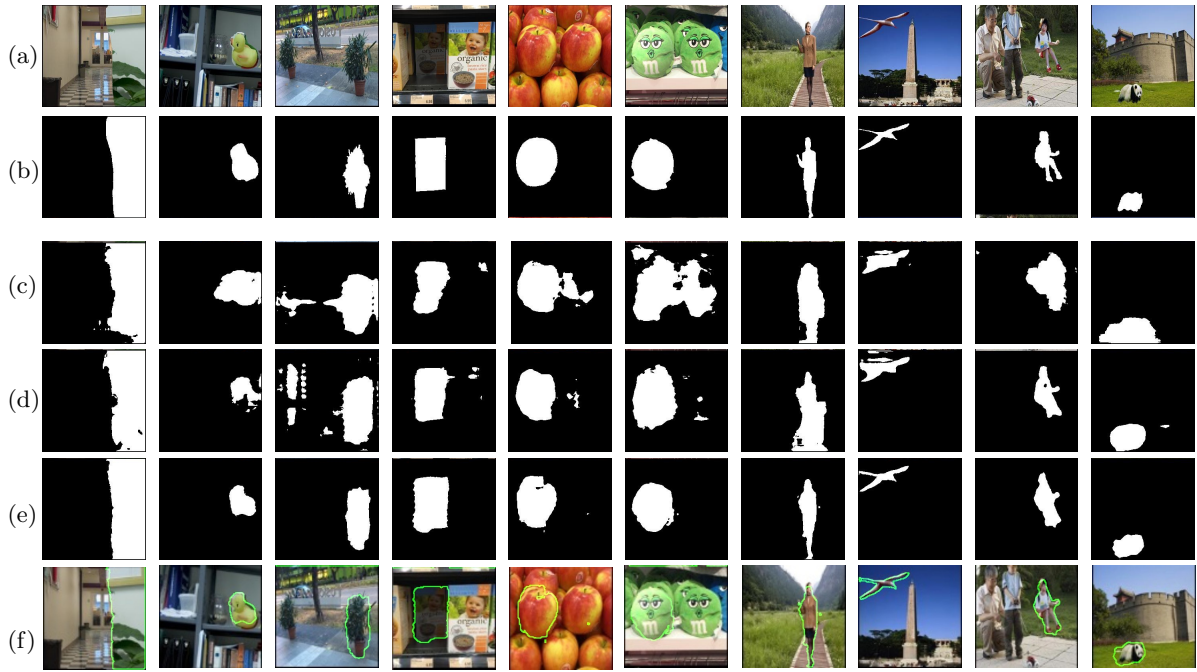
\begin{figure*}[t]
    \centering
    \begin{overpic}[width=0.95\textwidth]{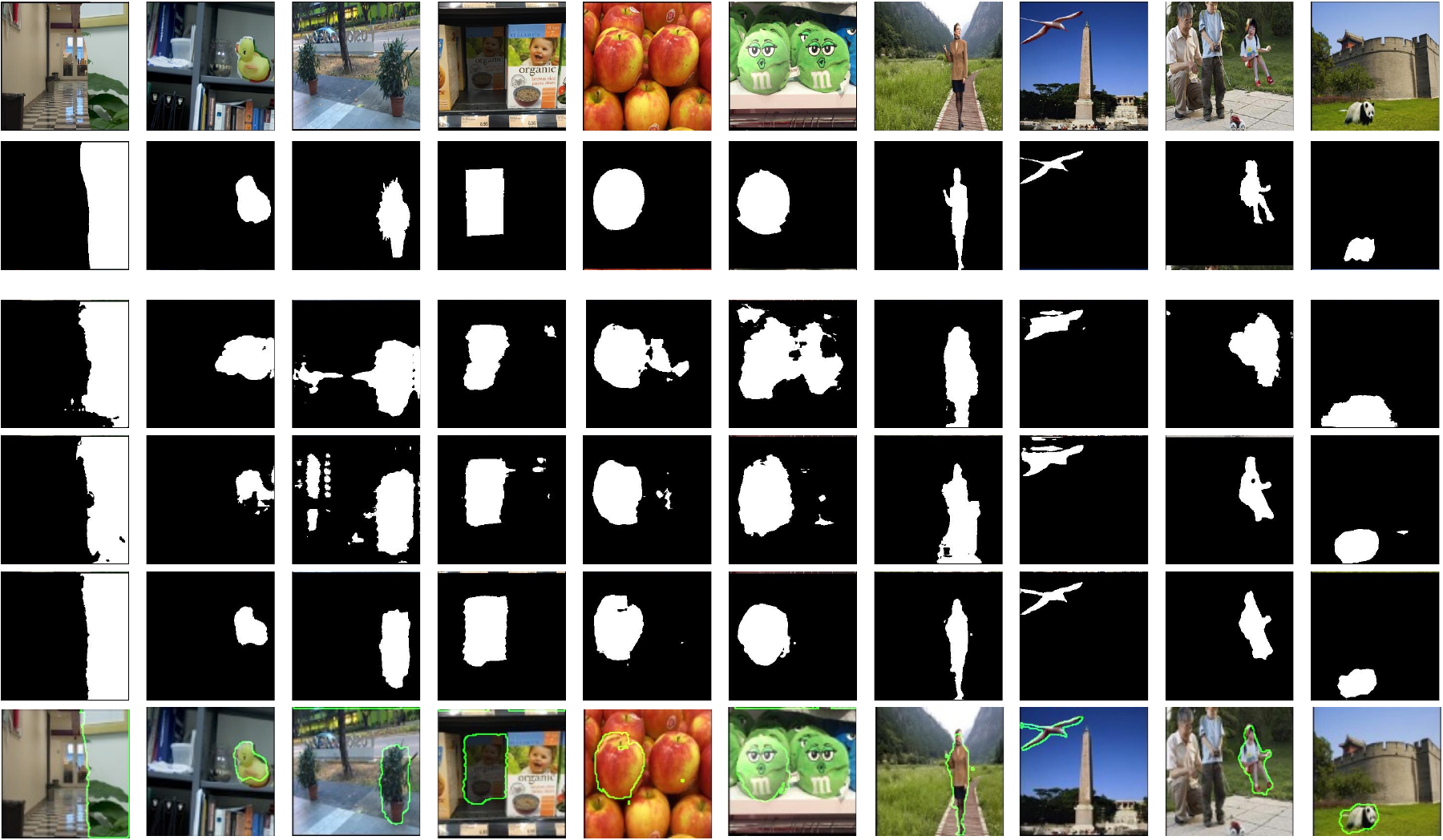}
    \put(-3,53){\small{(a)}}
    \put(-3,43){\small{(b)}}
    \put(-3,33){\small{(c)}}
    \put(-3,23){\small{(d)}}
    \put(-3,13){\small{(e)}}
    \put(-3,2){\small{(f)}}
    \end{overpic}
    \caption{Qualitative results on different image editing datasets. From top to bottom rows, these rows are (a) input image, (b) the ground truth mask, (c) PSCC-Net result, (d) HiFi-Net result~\cite{guo2023hierarchical}, (e) HiFi-Net++ result, and (f) the overlaid segmentation on the images of HiFi-Net++ result. \vspace{-3mm}}
    \label{fig_viz_manipulation}
\end{figure*}

\begin{table*}[t]
\begin{subtable}[b]{0.43\textwidth}
\centering
    \resizebox{1.\textwidth}{!}{
        \begin{tabular}{c|ccccc|c}
        \hline
        \multirow{2}{*}{Localization}
        &Col.&Cov.&NI.$16$&CAS.&IM$20$&Avg.\\ \cline{2-7}
        &\multicolumn{6}{c}{\textit{Metric:} AUC($\%$) -- Pre-trained} \\ \hline
        ManT.\cite{wu2019mantra} &$82.4$&$81.9$&$79.5$&$81.7$ &$74.8$&$80.0$\\
        SPAN\cite{hu2020span}&$93.6$&$92.2$&$84.0$&$79.7$&$75.0$&$84.9$\\ 
        PSCC-Net\cite{liu2022pscc}&$98.2$&$84.7$&$85.5$&$82.9$&$80.6$&$86.3$\\
        Ob.Fo.\cite{wang2022objectformer}&$95.5$&$92.8$&$87.2$&$84.3$&$82.1$&$88.3$\\
        HiFi-Net~\cite{guo2023hierarchical}&$98.3$&$\color{blue}93.2$& $87.0$&$85.8$&$82.9$&$89.4$\\
        NCL~\cite{ICCV_NCL}&$94.3$&$92.8$&$\mathbf{\color{red}{91.2}}$&$86.4$&$\color{blue}86.4$&$\color{blue}90.7$\\
        TANet~\cite{shi2023transformer}&$\mathbf{\color{red}{98.7}}$&$91.4$&$85.3$&$\mathbf{\color{red}{89.8}}$&$84.9$&$90.4$\\
        \hline
        HiFi-Net++ &$\color{blue}{98.5}$&$\mathbf{\color{red}{94.7}}$&$\color{blue}{87.4}$&$\color{blue}{88.6}$&$\mathbf{\color{red}{86.6}}$&$\mathbf{\color{red}{91.2}}$\\\hline
        \end{tabular}
    }
    \caption{}
\label{tab_img_editing_pre}
\end{subtable}\hfill
\begin{subtable}[b]{0.53\textwidth}
\centering
    \resizebox{1.\textwidth}{!}{
        \begin{tabular}{c| c c c | c}
        \hline
        \multirow{2}{*}{Localization}
        &Cov.&CAS.&NI.$16$&Avg.\\ \cline{2-5}
        &\multicolumn{4}{c}{\textit{Metric:} AUC($\%$) / F$1$($\%$) -- Fine-tuned} \\ \hline
        SPAN\cite{hu2020span}&$93.7/55.8$&$83.8/40.8$&$96.1/58.2$&$91.2/51.6$\\
        PSCC-Net\cite{liu2022pscc}&$94.1/72.3$&$87.5/55.4$&$99.6$/$81.9$&$93.7/69.8$\\
        Ob.Fo.\cite{wang2022objectformer}&$95.7$/$75.8$&$88.2$/$57.9$&$99.6$/$82.4$&$94.5$/$72.0$\\
        HiFi-Net~\cite{guo2023hierarchical} & $96.1$/$\color{blue}{80.1}$&$88.5$/$59.6$ & $98.9$/$85.0$&$94.6$/$\color{blue}75.5$\\
        TANet~\cite{shi2023transformer}&  $\mathbf{\color{red}{97.8}}$/$78.2$&$\color{blue}{89.3}$/$\color{blue}{61.4}$&$\mathbf{\color{red}99.7}$/$\color{blue}{86.5}$&$\color{blue}95.5$/$75.4$\\
        \hline
        HiFi-Net++ &$\color{blue}{97.5}$/$\mathbf{\color{red}{82.7}}$&$\color{red}\mathbf{90.3}$/$\mathbf{\color{red}65.4}$& $\mathbf{\color{red}99.7}$/$\mathbf{\color{red}86.8}$&$\mathbf{\color{red}95.8}$/$\mathbf{\color{red}76.6}$\\
        \hline
        \end{tabular}
    }
    \caption{}
\label{tab_img_editing_ft}
\end{subtable}\hfill
\caption{IFDL results on the image editing domain. (a) Localization performance of the pre-trained model. (b) Localization performance of the fine-tuned model. All results of prior works are ported from~\cite{wang2022objectformer}. [Key: \textcolor{red}{\textbf{Best}}; \textcolor{blue}{Second Best}. 
].}
\label{tab_img_editing}
\end{table*}
\begin{table*}[ht]
\centering
\begin{subtable}[b]{0.35\linewidth}
    \centering
    \begin{tabular}{c|cc}
        \hline
        Detection&AUC(\%)&F$1$(\%)\\ \hline
        ManT.\cite{wu2019mantra}&$59.9$&$56.7$\\
        SPAN\cite{hu2020span}&$67.3$&$63.8$\\
        PSCC-Net\cite{liu2022pscc}&$99.5$&$97.1$\\
        Ob.Fo.\cite{wang2022objectformer}&$\color{blue}99.7$&$97.3$\\
        \hline
        HiFi-Net\cite{guo2023hierarchical} & $99.5$&$\color{blue}{97.4}$\\  
        HiFi-Net++ & $\color{red}\mathbf{99.7}$&$\color{red}\mathbf{98.7}$\\ 
        \hline
    \end{tabular}
    \caption{}
    \label{tab_img_editing_img_level}
\end{subtable}\hfill
\begin{subtable}[b]{0.6\linewidth}
    \centering
    \resizebox{1.\textwidth}{!}{
    \begin{tabular}{c|ccccc}
    \hline    
        Post. &SPAN~\cite{hu2020span} &PSCC-Net~\cite{liu2022pscc} &Obj.Fo.~\cite{wang2022objectformer} & HiFi-Net~\cite{guo2023hierarchical} & HiFi-Net++\\ \hline
        
        Resize ($0.78$)&$83.24$&$85.29$&$87.2$&$86.9$&$\mathbf{87.7}$\\
        Resize ($0.25$)&$80.32$&$85.01$&$86.3$&$86.5$&$\mathbf{86.7}$\\ \hline
        
        Gau.Blur ($3$)&$83.10$&$85.38$&$85.97$&$86.1$&$\mathbf{86.9}$\\ 
        Gau.Blur ($15$)&$79.15$&$79.93$&$80.26$&$\mathbf{81.0}$&$80.1$\\ \hline
        
        Gau.Noi ($3$)&$75.17$&$78.42$&$79.58$&$\mathbf{81.9}$&$80.3$\\ 
        Gau.Noi ($15$)&$67.28$&$76.65$&$78.15$&$79.5$&$\mathbf{80.3}$\\ \hline
        
        J-Co. ($100$)&$83.59$&$85.40$&$86.37$&$86.3$&$\mathbf{87.4}$\\ 
        J-Co. ($50$)&$80.68$&$85.37$&$\mathbf{86.24}$&$86.0$&$85.6$\\ \hline\hline
        Aver.&$79.07$&$82.68$&$83.75$&$84.2$&$\textbf{85.1}$ \\ \bottomrule
    \end{tabular}}
    \caption{}
    \label{tab_editing_supp}
\end{subtable}\hfill
\caption{(a) Detection performance on \textit{CASIA} dataset. (b) Localization performance on \textit{NIST$\textit{16}$} with different post-processing steps. First three column results are ported from~\cite{wang2022objectformer}. Key: \textbf{Best}; Gau.: Gaussian; J-Co.: JPEG Compression.].}
\end{table*}

\begin{table*}[t]
\begin{subtable}{0.58\linewidth}
\centering
    \resizebox{1.\textwidth}{!}{
        \begin{tabular}{c|c|c|c|c|c}
            \toprule
            \multirow{2}{*}{Method} & \shortstack[c]{SD 1.5\\Inpaint} & \shortstack[c]{SD 2.1\\Inpaint} & \shortstack[c]{SD-XL\\Inpaint} & RePaint & Average \\ \cline{2-6}
            &\multicolumn{5}{c}{\textit{Metric:} AUC/F1($\%$)} \\ \hline
            PSCC-Net~\cite{liu2022pscc} & $.544/.395$ & $.509/.378$ & $.492/.436$ & $.632/.471$&$.511/.428$\\ \hline
            MVSS-Net~\cite{dong2022mvss} & $.683/.425$ & $.638/.471$ & $.606/.398$ & $.589/.416$&$.629/.414$\\ \hline
            HiFi-Net~\cite{guo2023hierarchical} & $\mathbf{.732}/.497$ & $.678/.483$ & $.685/.487$ & $.751/.559$&$.712/.506$\\ \hline
            HiFi-Net++ & $.728/\mathbf{.512}$ & $\mathbf{.723}/\mathbf{.568}$ & $\mathbf{.713}/\mathbf{.544}$ & $\mathbf{.754}/\mathbf{.578}$ & $\mathbf{.730}/\mathbf{.550}$\\
            \bottomrule
        \end{tabular}
    }
    \caption{}
\label{IJCV_zero_shot_localization}
\end{subtable}\hfill
\begin{subtable}{0.4\linewidth}
\centering
    \resizebox{1\textwidth}{!}{
        \begin{tabular}{c|c|c|c|c}
            \toprule
            \multirow{2}{*}{Method} & Backbone & Col. & Cov. & F-Shifter\\ \cline{2-5}
            &\multicolumn{3}{c}{\textit{Metric:} AUC($\%$)} \\ \hline
            \hline
            OneFormer~\cite{jain2023oneformer} & Swin-T &$.714$ & $.696$ & $.755$ \\ \hline
            OneFormer~\cite{jain2023oneformer} & ConvNext&$.723$ & $.707$ & $.805$ \\ \hline
            OneFormer~\cite{jain2023oneformer} & Dinat &$.686$ & $.652$ & $.774$ \\ \hline
            PSCC-Net~\cite{liu2022pscc} & HR-Net&$.787$ & $.739$ & $.831$ \\ \hline
            HiFi-Net++ & \shortstack[c]{Multi-branch\\Feature Extractor} &$\mathbf{.821}$ & $\mathbf{.786}$ & $\mathbf{.834}$ \\
            \bottomrule
        \end{tabular}
    }
    \caption{}
\label{tab:IJCV_oneformer_response}
\end{subtable}\hfill
\caption{Zero-shot localization performance on held-out GenAI methods and manipulation localization comparison with OneFormer~\cite{jain2023oneformer}. (a) The zero-shot localization performance on GenAI inpainting methods. (b) We apply OneFormer on the manipulation localization task and report the performance (AUC) in the comparison with PSCC-Net and HiFi-Net++. Different models' performances are evaluated on the HiFi-IFDL, and all methods are trained on $55,508$ images from the HiFi-IFDL dataset for a fair comparison. [Key: \textbf{Best}; SD: Stable Diffusion]. }
\label{IJCV_zero_shot}
\end{table*}

\SubSection{Image Editing Datasets Performance}\label{ex_data_editing}
Tab.~\ref{tab_img_editing} reports IFDL results for the image editing domain. We evaluate  on $5$ datasets: \textit{Columbia}~\cite{ng2009columbia}, \textit{Coverage}~\cite{wen2016coverage}, \textit{CASIA}~\cite{dong2013casia}, \textit{NIST$\textit{16}$}~\cite{NIST16} and \textit{IMD$20$}~\cite{IMD2020_wacv}. Following the previous experimental setup of~\cite{wu2019mantra,hu2020span,liu2022pscc,dong2022mvss,wang2022objectformer}, we pre-trained both HiFi-Net~\cite{guo2023hierarchical} and HiFi-Net++ on the HiFi-IFDL and then evaluate these two models on unseen test sets. 
After that, we continue to fine-tune the pre-trained models on the \textit{NIST$\textit{16}$}, \textit{Coverage}, and \textit{CASIA}. 
Tab.~\ref{tab_img_editing_pre} reports different pre-trained models' localization performances.
Specifically, NCL has a $1.3$ higher AUC score than HiFi-Net~\cite{guo2023hierarchical} on the average localization performance, showcasing the effectiveness of its contrastive learning scheme that captures the non-mutual exclusive relation among different image patches, TANet utilizes a stacked multi-scale transformer with boundary operators that help achieve $4.0$ higher AUC score than HiFi-Net~\cite{guo2023hierarchical} on CASIA dataset. 
Nevertheless, HiFi-Net++ maintains the best localization performance in this setup. For example, measured by the average performance across all different test sets, HiFi-Net++ has $0.5\%$ and $0.8\%$ higher AUC scores than TANet and NCL, respectively. This advantage largely comes from HiFi-Net++'s better performance on the Coverage~\cite{wen2016coverage} dataset, which contains many samples with only a small manipulation region. 
HiFi-Net++ is effective in identifying such manipulation areas with small spatial sizes. 
On the other hand, Tab.~\ref{tab_img_editing_ft} shows our HiFi-Net++ maintains the best localization performance when the model is fine-tuned on specific manipulation localization datasets, measured by either AUC or F1 score.
We hypothesize this is because the LGLE not only improves the HiFi-Net++'s localization performance to unseen manipulation patterns but also encourages HiFi-Net++ to capture specific manipulation patterns after fine-tuning.

\Paragraph{Image-level Forgery Detection} We also report the image-level forgery detection results in Tab.~\ref{tab_img_editing_img_level}, in which HiFi-Net++ achieves the best performance, while HiFi-Net achieves comparable results to ObjectFormer~\cite{wang2022objectformer}. This shows the language-guided forgery localization enhancer also yields an effective representation that can potentially help detect the image manipulated by the conventional editing method. We show qualitative results in Fig.~\ref{fig_viz_manipulation}, where the manipulated region identified by HiFi-Net++ can capture semantically meaningful object shapes (\textit{e.g.}, person and airplane) and offer much more accurate localization performance than HiFi-Net and PSCC-Net.

\Paragraph{Robustness Analysis} Following previous works~\cite{wang2022objectformer,liu2022pscc,hu2020span}, we evaluate the robustness of both HiFi-Net and HiFi-Net++ against different post-processing steps. The corresponding result is reported in Tab.~\ref{tab_editing_supp}. First, the proposed HiFi-Net is more robust than the previous work, except for the post-processing of resizing $0.78$ times the image and JPEG compression with $50\%$ quality. On the other hand, the robustness further improves in the HiFi-Net++, with an increase of $0.9\%$ in the average AUC over HiFi-Net. This improvement can be explained by the fact that HiFi-Net++ uses the visual embedding of the pre-trained CLIP image encoder. This visual embedding is exposed to data collected online, which contains a large number of noisy and blurred images.

\Paragraph{Zero-shot Localization Performance on GenAI Inpainting Methods} 
To verify the effectiveness of the proposed localization methods on the image manipulated by recent GenAI methods, we collect images manipulated by Stable Diffusion Models and Repaint~\cite{lugmayr2022repaint}, which are held out from training, thereby testing HiFi-Net++ zero-shot localization performance. 
Tab.~\ref{IJCV_zero_shot_localization} reports the zero-shot localization performance on $4$ GenAI inpainting methods.
Our proposed HiFi-Net++ achieves the best localization performance among the compared methods, except in the case of Stable Diffusion 1.5 Inpainting, where HiFi-Net++ has a lower AUC score than HiFi-Net~\cite{guo2023hierarchical}.

\begin{table*}[t]
  \centering
    \resizebox{1\linewidth}{!}{
    \begin{tabular}{c ccccc c cc cc ccccccc cc}
        \toprule
        \multirow{2}{*}{\shortstack[c]{Detection\\method}} & \multicolumn{6}{c}{Generative Adversarial Networks} & \multirow{2}{*}{\shortstack[c]{Deep-\\fakes}} & \multicolumn{2}{c}{Low level vision} & \multicolumn{2}{c}{Perceptual loss} & \multicolumn{7}{c}{Gen-AI} & \multirow{2}{*}{Avg.} \\
        \cmidrule(lr){2-7} \cmidrule(lr){10-11} \cmidrule(lr){11-12} \cmidrule(lr){13-19} 
        & \shortstack[c]{Pro-\\GAN} & \shortstack[c]{Cycle-\\GAN} & \shortstack[c]{Big-\\GAN} & \shortstack[c]{Style-\\GAN} & \shortstack[c]{Gau-\\GAN} &  \shortstack[c]{Star-\\GAN} & FF++ & SITD & SAN & CRN & IMLE & \shortstack[c]{DALL\\E-3} & \shortstack[c]{Mid\\journey} & \shortstack[c]{SD\\2.1} & \shortstack[c]{SD\\XL} & IF & SORA & \shortstack[c]{Instant\\ID} &\\ 
        \midrule
        DE-FAKE~\cite{sha2023de-fake} & $96.03$ & $97.74$ & $89.93$ & $78.09$ & $98.15$ & $98.34$ & $81.43$ & $88.71$ & $\mathbf{\textcolor{red}{90.14}}$ & $90,01$ & $95,14$ & \textcolor{red}{$\mathbf{88.75}$} & $66.60$ & $79.28$ & $40.65$ & $66.61$ & $27.43$ & $50.11$ & $79.06$\\
        \midrule
        Uni-Det.~\cite{ojha2023towards} & $\mathbf{\textcolor{red}{100.0}}$ & $\mathbf{\textcolor{red}{99.46}}$ & $\mathbf{\textcolor{red}{99.59}}$ & $97.24$ & $\mathbf{\textcolor{red}{99.98}}$ & $99.60$ & $82.45$& $61.32$ & $79.02$ & $91.72$ & $99.00$ & $74.53$ & $64.93$ & \textcolor{red}{$\mathbf{95.29}$} & $86.37$ & $90.26$ & $67.94$ & $34.62$ & $84.91$ \\
        \midrule
        DIRE~\cite{wang2023dire}&$53.63$ & $51.63$ & $49.11$ & $34.39$ & $48.44$ & $59.23$ & $48.74$ & $48.52$ & $53.13$ & $49.96$ & $50.22$& $74.70$ & $\textcolor{red}{\mathbf{99.99}}$ & $47.60$ & $49.27$ & $\mathbf{\textcolor{red}{100.0}}$ & $87.62$ & $\mathbf{\textcolor{red}{100.0}}$ & $61.45$\\
        \midrule
        Ours & $\mathbf{\textcolor{red}{100.0}}$ & $98.54$ & $98.48$ & $\mathbf{\textcolor{red}{99.60}}$ & $97.19$ & $\mathbf{\textcolor{red}{99.99}}$ & $\mathbf{\textcolor{red}{83.28}}$ & $\mathbf{\textcolor{red}{96.11}}$ & $81.57$ & $\mathbf{\textcolor{red}{92.41}}$&$\mathbf{\textcolor{red}{99.96}}$ & $59.31$ & $78.32$ & $89.00$ & \textcolor{red}{$\mathbf{92.35}$} & $84.59$ & \textcolor{red}{$\mathbf{93.24}$} & $72.80$ & $\mathbf{\textcolor{red}{89.71}}$\\
        \bottomrule
    \end{tabular}
}
\vspace{1.5mm}
\caption{Image forgery detection performance measured by Average Precision. Forgery methods include $11$ generative methods used in previous work~\cite{wang2020cnn,ojha2023towards} and $7$ representative Gen-AI methods. Previous methods' quantitative results are taken from \cite{ojha2023towards}. 
Each prior detections methods have many variants, so we select the one with the best forgery detection performance for the comparison. 
[\textbf{\textcolor{red}{Bold}}: best result.]}
\label{tab:IJCV_detection_response}
\end{table*}

\Paragraph{Localization comparison with OneFormer}
OneFormer~\cite{jain2023oneformer}, a recently proposed segmentation method, exhibits strong performance in dense prediction tasks. Therefore, we utilize OneFormer for the manipulation localization task and present the results in Table \ref{tab:IJCV_oneformer_response} to provide a more comprehensive comparative study. The best OneFormer variant is based on ConvNeXt and achieves a worse localization performance compared to PSCC-Net: $0.64$ and $0.32$ lower AUC scores on Columbia and Coverage, respectively. 
We believe the reason for OneFormer's poor manipulation localization performance is that OneFormer is a segmentation method, and there is an inherent difference between segmentation and manipulation localization tasks. As discussed in the previous work~\cite{dong2022mvss,zhou2020generate}, the segmentation requires the proposed method to have abilities in object recognition and semantics understanding, yet manipulation localization focuses on learning subtle visual artifacts and frequency domain differences among real and forged pixels, which can be almost invisible to humans.

\SubSection{CNN Image Detection}
\label{ex_cnn_det}
Tab.~\ref{tab:IJCV_detection_response} reports different methods' detection performances on CNN-generated images. Specifically, consistent with Uni-Det, we use the identical training dataset, which contains forged images generated from ProGAN~\cite{karras2018progressive} and real images from LSUN dataset~\cite{yu2015lsun}. We employ Average Precision (AP) as the metric, the same as the previous work, for a fair comparison. To have a thorough evaluation, we construct a forgery detection benchmark including $11$ generative methods that are commonly used for measuring generalizable detection performance~\cite{wang2020cnn,jeong2022frepgan,jeong2022bihpf,ricker2022towards,ojha2023towards} and $7$ representative GenAI methods. More formally, to have better diversity in forgery methods, these GenAI methods span from classical Stable Diffusion models~\cite{rombach2021highresolution_latent_diffusion}, commercial tools (\textit{e.g.}, Sora, Midjourney, and DALLE-3), to the recent popular personalized method like InstantID~\cite{wang2024instantid}.
Also, following the previous work~\cite{wang2020cnn}, we add real images that have identical semantics (\textit{i.e.}, face and non-face objects) as forged images in this benchmark.

As indicated by Tab.~\ref{tab:IJCV_detection_response}, our proposed HiFi-Net++ achieves the best image forgery detection performance, surpassing the second and third best detection methods (\textit{i.e.}, Uni-Det and DE-FAKE) by $4.8\%$ and $10.65\%$ in terms of Average Precision, respectively.
Although Uni-Det and DE-FAKE leverage CLIP pre-trained encoders with powerful visual representations that help detect forged images, they only achieve the best detection performance on $5$ and $2$ individual forgery methods, respectively. In contrast, HiFi-Net++ achieves the best detection performance on $9$ individual forgery method detections.
We believe this performance gap is due to two factors. First, HiFi-Net's multi-branch feature extractor is carefully devised and can learn more comprehensive artifacts from generative images, improving the overall image-level forgery detection performance. Secondly, our HiFi-Net++ is trained for both forgery detection and manipulation localization tasks, and such multi-task learning further enhances the representation's effectiveness in capturing forgery attributes. 
Additionally, DIRE achieves the best performance over images generated from GenAI methods ($3$ best performance out of $7$ Gen-AI methods), whereas it suffers from poor detection precision among images generated from GANs (\textit{e.g.}, ProGAN, CycleGAN, etc.). This phenomenon indicates that DIRE's detection performance is limited to certain forgery types, and its overall performance is worse than HiFi-Net++.

However, it is worth mentioning that HiFi-Net++ detection performance on DALLE-3 images falls behind prior arts. 
This is because the multi-branch feature extractor in the HiFi-Net++, which helps achieve leading average detection AP, the most important aspect for real-world applications, might be less effective in detecting DALLE-3.
However, a simple modification --- reducing the number of branches in HiFi-Net++ --- significantly improves detection AP on DALLE-3.
Tab.~\ref{tab:rebuttal_ijcv} reports the 1-branch and 2-branch HiFi-Net++ variants, which achieve $24.7\%$ and $11.9\%$ higher AP than the 4-branch HiFi-Net++ variant (\textit{i.e.}, full model), respectively.
Nevertheless, these HiFi-Net++ variants with fewer branches ineffectively learn diverse forgery patterns, decreasing averaged detection performance. 
This suggests that a meaningful future research direction is to develop an adaptive branch control mechanism, dynamically adjusting the number of branches based on the input image for better overall performance.

\begin{table}[t]
    \centering
    \scalebox{1}{
        \centering
        \begin{tabular}{c|c|c}
            \toprule
            \multirow{2}{*}{\shortstack[c]{HiFi-Net++\\variants}} & DALLE-3 & Avg. \\ \cline{2-3}
            &\multicolumn{2}{c}{\textit{Metric:} AP($\%$)} \\ \hline
            $1$-branch & $\mathbf{84.0}$ & $78.1$ \\
            $2$-branch & $71.2$ & $83.1$ \\
            $4$-branch (\textit{full model}) & $59.3$ & $\mathbf{89.7}$ \\
            \bottomrule
        \end{tabular}
    }
    \caption{
    \textcolor{black}{Detection performance of HiFi-Net++ variants on DALLE-3 images and their averaged performance across $18$ forged methods in Tab.~\ref{tab:IJCV_detection_response}. [Key: \textbf{Best}, Avg.: Average detection performance, AP: Average Precision.].}
    }
    \label{tab:rebuttal_ijcv}
\end{table}

\SubSection{Diverse Fake Face Dataset Performance}\label{ex_data_dffd}

\begin{table}[t]
\footnotesize
\begin{subtable}{1.\linewidth}
    \centering
    \resizebox{1\textwidth}{!}{
    \begin{tabular}{c|c c c}
        \hline
        IoU ($\uparrow$) / PBCA ($\uparrow$)& Real & Fu. Syn. & Par. Man.\\
        \hline
        Att.~\cite{stehouwer2019detection} &$-/\mathbf{0.998}$&$0.847/0.847$&$0.401/0.786$\\\hline
        HiFi-Net~\cite{guo2023hierarchical} 
        &$-/0.978$&$0.893/0.893$&$0.411/0.801$ \\  \hline
        HiFi-Net++ 
        &$-/0.974$&$\mathbf{0.901}/\mathbf{0.901}$&$\mathbf{0.412}/\mathbf{0.830}$ \\ \hline
    \end{tabular}}
    \caption{Localization performance measured by IoU and PBCA, which are the higher, the better.}
    \label{tab_dffd_loc}
\end{subtable}
    \centering
\begin{subtable}{1.\linewidth}
    \centering
    \resizebox{1\textwidth}{!}{
    \begin{tabular}{c|c c c}
        \hline
        IINC ($\downarrow$) / C.S. ($\downarrow$)& Real & Fu. Syn. & Par. Man.\\
        \hline
        Att.~\cite{stehouwer2019detection} &$0.015/-$&$0.077/0.095$&$\mathbf{0.311}/0.429$\\\hline
        HiFi-Net~\cite{guo2023hierarchical} 
        &$\mathbf{0.010}/-$&$0.060/0.107$&$0.323/0.410$ \\  \hline
        HiFi-Net++&$0.012/-$&$\mathbf{0.057}/\mathbf{0.092}$&$0.313/\mathbf{0.402}$ \\  \hline
    \end{tabular}}
    \caption{Localization performance measured by IINC and Cosine Similarity, which are the lower, the better.}
    \label{tab_dffd_loc_2}
\end{subtable}
\begin{subtable}{0.5\linewidth}
    \centering
    \resizebox{1\textwidth}{!}{
        \begin{tabular}{c|c}
        \hline
        \multicolumn{2}{c}{\textit{Metric:} AUC} \\
        \hline
        Att.Xce.~\cite{stehouwer2019detection} &$99.69$\\
        \hline
        HiFi-Net~\cite{guo2023hierarchical} & $99.45$ \\  \hline
        HiFi-Net++ & $\mathbf{99.71}$ \\  \hline
        \end{tabular}
    }
    \caption{Detection Performance.}
    \label{tab_dffd}
\end{subtable}
\caption{IFDL results on DFFD dataset. [Keys: Key: \textbf{Best}. Fu. Syn.: Fully-synthesized; Par. Man.: Partially-manipulated. C.S.: consine similarity].}
\end{table}

For the facial image forgery domain, we evaluate our methods on the Diverse Fake Face Dataset (DFFD)~\cite{stehouwer2019detection}, which has fake facial images synthesized by different facial forgery methods and real faces from FFHQ~\cite{karras2019style} as well as CelebA~\cite{liu2015faceattributes_celeba}.
For a fair comparison, we follow the same experiment setup and metrics: for the pixel-level localization, we use IoU, Cosine Similarity, pixel-wise binary classification accuracy (PBCA), and IINC defined work~\cite{stehouwer2019detection}; for image-level detection, we use AUC to measure. First, Tab.~\ref{tab_dffd_loc} and Tab.~\ref{tab_dffd_loc_2} show that HiFi-Net++ can produce better localization performance than the previous work. 
Specifically, HiFi-Net++ achieves the best localization on fully synthesized fake facial images ($0.901$ IOU and $0.057$ IINC). 
As for HiFi-Net, it also generates a localization area that has higher IOU and PBCA than the previous work on partially manipulated and fully synthesized images. 
Secondly, Tab.~\ref{tab_dffd} reports that, on the detection, both HiFi-Net++ and HiFi-Net achieve comparable performance with the previous method. 
This is still an impressive detection result since the prior work~\cite{stehouwer2019detection} uses a more powerful backbone architecture and is specific to the digital facial forgery domain.

Lastly, it is an interesting research direction that explores trained deepfake detection algorithms' performance in the physical domain, \textit{i.e.}, face anti-spoofing. 
For example, it has been discussed in the previous work~\cite{guo2023hierarchical} that some spoof images from SiW-Mv2~\cite{guo2022multi} are identified as digital forget images.

\subsection{Ablation Study}\label{ex_ablation}
\begin{table}[t]
    \begin{subtable}{1\linewidth}
        \centering
        \footnotesize
        \resizebox{1\linewidth}{!}{
        \begin{tabular}{ccc|cc|cc}
            \hline
            \multirow{2}{*}{}&\multirow{2}{*}{Modules}&\multirow{2}{*}{Loss}
            &\multicolumn{2}{c|}{Detection}&\multicolumn{2}{c}{Localization}\\ \cline{4-7}
            &&&AUC &F$1$ &AUC &F$1$\\ \hline\hline

            \textit{1}&\textbf{L},\textbf{C}, \textbf{LFLE}&$\mathcal{L}_{cls}$, $\mathcal{L}_{loc}$, $\mathcal{L}^{score}_{loc}$&$\mathbf{97.0}$&$\mathbf{94.3}$&$\mathbf{97.7}$&$\mathbf{98.6}$\\ \hline
            \textit{2}&\textbf{L},\textbf{C}, \textbf{LFLE}&$\mathcal{L}_{cls}$, $\mathcal{L}_{loc}$&$96.6$&$93.9$&$95.7$&$97.4$\\ \hline
            \textit{3}&\textbf{L},\textbf{C}&$\mathcal{L}_{cls}$, $\mathcal{L}_{loc}$&$96.8$&$94.1$&$95.3$&$96.9$\\ \hline
            \textit{4}&\textbf{L}&$\mathcal{L}_{loc}$&$65.0$&$70.0$&$93.4$&$95.0$\\ \hline
            \textit{5}&\textbf{C}&$\mathcal{L}_{cls}$&$95.8$&$92.4$&$66.0$&$58.0$\\ \hline
            \textit{6}&\textbf{L},\textbf{C}&$\mathcal{L}^{4}_{cls}$,$\mathcal{L}_{loc}$&$93.1$&$91.7$&$92.5$&$93.9$\\ \hline
            \textit{7}&\textbf{L},\textbf{C}&$\mathcal{L}^{ind}_{cls}$,$\mathcal{L}_{loc}$&$93.2$&$92.8$&$93.2$&$94.8$\\ \hline
        \end{tabular}}
    \caption{Ablation study on the Classification module, Localization modules, and Language-guided Forgery Localization Enhancer, denoted as \textbf{C}, \textbf{L}, and \textbf{LFLE}, respectively.
    $\mathcal{L}_{cls}$, $\mathcal{L}_{loc}$, and $\mathcal{L}^{score}_{loc}$ are loss functions for classification, localization, and manipulation score maps, respectively. $\mathcal{L}^{4}_{cls}$ and $\mathcal{L}^{ind}_{cls}$ denote we only perform the fine-grained classification on $4^{th}$ level and classification without hierarchical path prediction. These model variants' performances are based on the image feature extracted by the proposed multi-branch feature extractor, which captures comprehensive generation artifacts.}
    \label{table_ablation}
    \end{subtable}
    \begin{subtable}{1\linewidth}{
        \footnotesize
        \centering
        \resizebox{\linewidth}{!}{
        \begin{tabular}{cccc|cccc}
            \toprule
            Local.&\multirow{2}{*}{\begin{tabular}[c]{@{}c@{}}Image\\Encoder\end{tabular}}&\multirow{2}{*}{\begin{tabular}[c]{@{}c@{}}Pre-\\training\end{tabular}}&\multirow{2}{*}{\begin{tabular}[c]{@{}c@{}}Mani.\\Scr. Mask\end{tabular}}
            &\multicolumn{2}{c}{Image Edit.}&\multicolumn{2}{c}{CNN-syn.}\\ \cline{5-8}
            
            &&&&AUC &F$1$&AUC &F$1$\\ \midrule
            
            \textit{1}&ResNet-50& CLIP~\cite{radford2021learning}& \checkmark &$\mathbf{97.8}$&$\mathbf{94.7}$&$98.6$&$\mathbf{98.1}$\\\hline
            \textit{2}&ResNet-50 &CLIP~\cite{radford2021learning}& &$96.0$&$93.6$&$\mathbf{98.9}$&$97.6$\\
            \textit{3}&ResNet-101&CLIP~\cite{radford2021learning}& &$95.6$&$93.3$&$98.1$&$98.1$\\
            \textit{4}&ViT-B-16&CLIP~\cite{radford2021learning}& &$95.8$&$93.8$&$98.1$&$97.4$\\\hline
            \textit{5}&ResNet-50$^{1}$&ImageNet~\cite{deng2009imagenet}& &$74.0$&$56.2$&$87.0$&$70.5$\\
            \textit{6}&HR-Net$^{2}$&Cityscapes~\cite{cordts2016cityscapes}& &$89.8$&$77.6$&$93.4$&$79.5$\\\hline
            \textit{Baseline}&---&---& &$93.0$&$90.1$&$98.4$&$97.0$\\ \bottomrule
        \end{tabular}}
    \caption{Ablation study on alternative designs of language-guided forgery localization enhancer. ResNet-50$^{1}$ is pre-trained on ImageNet dataset for image recognition. HR-Net$^{2}$~\cite{wang2020deepHRNet} is pre-trained on Cityscapes dataset for the semantic segmentation.}
    \label{table_ablation_clip}
    }\end{subtable}
\caption{Ablation study on HiFi-Net++. [Key: \textbf{Best}, Mani. Scr. Mask: manipulation score masks].}
\end{table}

To provide a more detailed view of how each proposed component contributes to the final performance, we first report the ablation study on localization and classification module in Tab.~\ref{table_ablation}, and then we use Tab.~\ref{table_ablation_clip} to report the ablation study on the language-guided forgery localization enhancer.

\Paragraph{HiFi-Net++ Architecture} We start with row~$1$ in Tab.~\ref{table_ablation}, which represents the performance of the full HiFi-Net++, supervised by $\mathcal{L}_{loc}$, $\mathcal{L}_{cls}$, and $\mathcal{L}^{score}_{loc}$. 
From row~$1$ to $2$, we first ablate the $\mathcal{L}^{score}_{loc}$, which reduces both detection and localization performances. This is because it is important to have $\mathcal{L}^{score}_{loc}$, which helps Language-guided Forgery Localization Enhancer (\textit{i.e.}, LFLE) generate accurate manipulation score maps that serve as the spatial guidance for identifying the manipulated region.
Also, comparing row~$1$ with row~$3$, we can see LFLE improves the localization performance ($2.4\%$ AUC and $1.7\%$ F1 score), which is consistent with the statement that LFLE improves the effectiveness of the proposed method in identifying manipulated pixels.
Furthermore, compared to row~$3$'s performance, we ablate the classification module and localization module in row~$4$ and $5$, removing which causes large performance drops on the detection ($24.1\%$ F$1$) and localization ($29.3\%$ AUC), respectively. 
We evaluate the effectiveness of performing fine-grained classification at different hierarchical levels. 
In the $6$th row, we only keep the $4$th level fine-grained classification in training, leading to a sensible drop of performance in detection ($3.7\%$ AUC) and localization ($2.8\%$ AUC). 
In the $7$th row, we perform the fine-grained classification without forcing the hierarchy of~Eq.~\ref{eq_conditional_prob}, which decreases $3.6\%$ AUC in the detection.

\Paragraph{Language-guided Forgery Localization Enhancer} In Tab.~\ref{table_ablation_clip}, row $1$ denotes the HiFi-Net++ with a complete language-guided forgery localization enhancer, which has the pre-trained image encoder and uses manipulation score maps as the auxiliary signals. First, from row $1$ to row $2$, we ablate the manipulation score map generated by text inputs. As a result, the localization performance drops $1.8\%$ AUC and $0.9\%$ F$1$ scores in the image editing domain. Although the localization performance is not affected largely in the CNN-synthesis domain, we still believe that generated manipulation score maps help the overall localization performance. This phenomenon further shows the pre-trained CLIP model's ability to associate the input text with the corresponding visual contents. After that, we experiment with different image encoder backbones (from row $2$ to $4$), such as ResNet$50$, ResNet$100$ and ViT-B-$16$. Apparently, all these image encoder backbones provide a robust image embedding that enables the distinction between real and manipulated pixels, and these performances do not vary largely from each other. Subsequently, we replace the pre-trained CLIP image encoder with different ones that are trained for different tasks. Specifically, the performance gap between row $2$ and row $5$, as well as row $6$, indicates the fact that visual embedding from the pre-trained CLIP image encoder is the better feature space for the manipulation localization task. In the last row, comparing to this baseline performance, we can observe that regardless of which pre-trained CLIP image encoder is used, the model always achieves the better localization performance over the baseline. For example, when using ResNet$50$ as the image encoder backbone, the localization performance improves $3.0\%$ AUC and $3.5\%$ F$1$ in the image editing domain, and $0.5\%$ higher AUC in the CNN-synthesis domain.

\Paragraph{Hierarchical Fine-grained Scheme} To show how different hierarchical fine-grained schemes (HiFi-schemes) contribute to the IFDL performance, we compare our proposed Hierarchical Path Prediction (\textit{e.g.}, $\mathcal{L}_{path}$) in HiFi-Net++ with the HiFi-scheme proposed in HRN~\cite{chen2022label} and evaluate their performances on the HiFi-IFDL dataset. Specifically, HRN's HiFi-scheme employs a probabilistic classification $\mathcal{L}_{Hier.}$ with a pre-defined state space $S$ to help capture the hierarchy among different tree nodes. Tab.~\ref{tab_IJCV_HRN_performance} reports IFDL performance of two HiFi-schemes. The line~$\#1$ indicates the performance without the HiFi-scheme, while the line~$\#2$ and line~$\#3$ indicate performances using HiFi-schemes from HRN (\textit{i.e.}, state space $S$ with $\mathcal{L}_{Hier.}$) and HiFi-Net++ (\textit{i.e.}, $\mathcal{L}_{path}$), respectively. 
Comparing the performance of line~$\#1$ against line~$\#2$ and $\#3$, we can conclude that the HiFi-scheme contributes to IFDL performance. Moreover, using the $\mathbf{L}_{path}$ can achieve slightly better localization performance and comparable detection performance as HRN's $\mathbf{L}_{hier.}$ and $S_{F}$. This indicates the effectiveness of our proposed $\mathbf{L}_{path}$ in capturing the inherent hierarchical correlation among various forgery attributes.

\begin{table}[t]
    \centering
    \footnotesize
    \resizebox{1\linewidth}{!}{
    \begin{tabular}{c|c|cc|cc}
        \hline
        &\multirow{2}{*}{Method}&\multicolumn{2}{c|}{Detection}&\multicolumn{2}{c}{Localization}\\ \cline{3-6}
        &&AUC &F1 &AUC &F1\\ \hline\hline
        \textit{1}&-&$93.2$&$92.8$&$93.2$&$94.8$\\ \hline
        \textit{2}&$S$,$\mathcal{L}_{hier.}$~\cite{chen2022label}&$95.5$&$\mathbf{94.2}$&$94.9$&$95.9$\\ \hline
        \textit{3}&$\mathcal{L}_{path}$ (ours)&$\mathbf{96.8}$&$94.1$&$\mathbf{95.3}$&$\mathbf{96.9}$\\ \hline
    \end{tabular}
    }
    \vspace{1.5mm}
    \caption{IFDL performance of HiFi-Net++ using different hierarchical fine-grained schemes. [Key: \textbf{Bold}.]}
    \label{tab_IJCV_HRN_performance}
\end{table}

\begin{figure}[t!]
    \centering
    \includegraphics[width=0.45\textwidth]{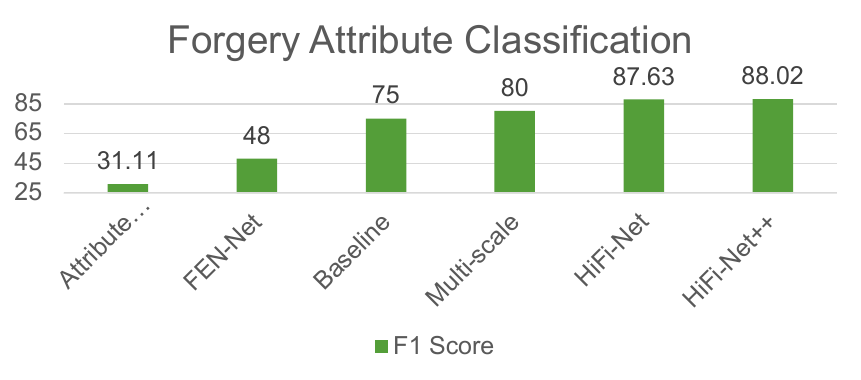}
    \caption{The forgery attribute classification results.}
    \label{tab_fg_performance}
\end{figure}

\SubSection{Forgery Attribute performance}\label{ex_mani_attribute}
We perform the fine-grained classification among real images and $13$ forgery categories on $4$ different levels, and the most challenging scenario is the fine-grained classification on the $4$-th level.
The result is reported in Fig.~\ref{tab_fg_performance}. Specifically, we train HiFi-Net $4$ times, and at each time, only classifies the fine-grained forgery attributes at one level, denoted as \textit{Baseline}. 
Then, we train a HiFi-Net~\cite{guo2023hierarchical} to classify all $4$ levels but without the hierarchical dependency via Eq.~\ref{eq_conditional_prob}, denoted as \textit{multi-scale}. 
Also, we compare the pre-trained image attribution works~\cite{asnani2021reverse,yu2019attributing_image_attribute} and the new proposed HiFi-Net++. Both HiFi-Net and HiFi-Net++ have improvements over previous works~\cite{yu2019attributing_image_attribute,asnani2021reverse}, which we believe is because the previous works only learn to attribute CNN-synthesized images, yet do not consider attributing image editing methods. 
Again, the performance difference between HiFi-Net and HiFi-Net++ is not large since the language-guided forgery localization enhancer mainly improves the localization performance.
\Section{Conclusion}
In this work, we develop an IFDL method for both CNN-synthesized and image-editing forgery domains.
We formulate the IFDL as a hierarchical fine-grained classification problem that requires the algorithm to classify the individual forgery method of given images via predicting the entire hierarchical path. Moreover, we improve our algorithm generalization ability and performance on localizing small-region manipulations via a language-guided forgery localization enhancer. Lastly, HiFi-IFDL dataset is proposed to further help the community in developing forgery detection algorithms.

\Paragraph{Data Availability Statement}
The Hierarchical Fine-grained IFDL dataset---HiFi-IFDL dataset---is publicly available at the following repository \href{https://github.com/CHELSEA234/HiFi_IFDL}{github.com/CHELSEA234/HiFi-IFDL}.

\Paragraph{Acknowledgments}
This work was supported by the following projects. One is the Defense Advanced Research Projects Agency (DARPA) under Agreement No. HR$00112090131$ at Michigan State University, the others are Sapienza research projects ``Prebunking'', ``Adagio'', and ``Risk and Resilience factors in disadvantaged young people: a multi-method study in ecological and virtual environments''.

\bibliographystyle{sn-basic} 
\small{
\bibliography{dfd}
}

\end{document}